\documentclass{article}

\usepackage{arxiv}

\usepackage[utf8]{inputenc} 
\usepackage[T1]{fontenc}    
\usepackage{hyperref}       
\hypersetup{
    colorlinks,
    citecolor=blue,
    filecolor=black,
    linkcolor=blue,
    urlcolor=black
}
\usepackage{url}            
\usepackage{booktabs}       
\usepackage{amsmath,amssymb,amsfonts}       
\usepackage{nicefrac}       
\usepackage{microtype}      
\usepackage{cleveref}       
\usepackage{lipsum}         
\usepackage{graphicx}
\usepackage{doi}

\usepackage{IEEEtrantools}
\usepackage{soul}
\usepackage{cite}
\usepackage{float}
\usepackage{caption}
\usepackage{subcaption}
\usepackage{stfloats}
\usepackage{enumitem}
\usepackage{mathtools}
\usepackage{algorithm}
\usepackage{algpseudocode}
\usepackage{graphicx,color}
\usepackage{textcomp}
\usepackage{xcolor}
\usepackage{bm}
\usepackage{diagbox}
\usepackage{multirow}
\usepackage{makecell}
\usepackage{cases}
\usepackage{tikz}
\usepackage[british]{babel}
\usepackage[mathscr]{eucal}
\newcommand\norm[1]{\left\lVert#1\right\rVert}

\title{On the Limitations of Physics-informed Deep Learning: Illustrations Using First Order Hyperbolic Conservation Law-based Traffic Flow Models}

\date{}

\author{
    \href{https://orcid.org/0000-0001-6736-5627}{\includegraphics[scale=0.06]{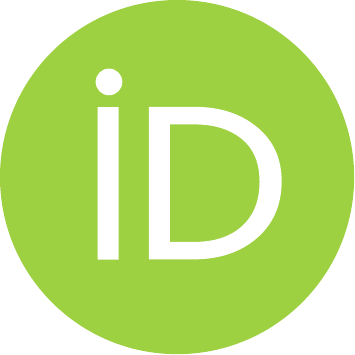}\hspace{1mm}Archie J.~Huang}\,,
    \href{https://orcid.org/0000-0001-7754-6341}{\includegraphics[scale=0.06]{figures/orcid.pdf}\hspace{1mm}Shaurya~Agarwal}\\
    UrbanITY Lab\thanks{The code accompanying this manuscript can be found in the GitHub repository at \href{https://github.com/Urbanity-Lab/PIDL-Limitations}{github.com/Urbanity-Lab/PIDL-Limitations}}\\
    Civil, Environmental \& Construction Engineering Department\\
    University of Central Florida\\
    Orlando, FL 32816, USA \\
    \texttt{archie.huang@nyu.edu, shaurya.agarwal@ucf.edu} \\
}

\renewcommand{\shorttitle}{On the Limitations of Physics-informed Deep Learning}

\hypersetup{
pdftitle={On the Limitations of Physics-informed Deep Learning},
pdfsubject={Illustrations Using First Order Hyperbolic Conservation Law-based Traffic Flow Models},
pdfauthor={Archie J.~Huang, Shaurya~Agarwal},
pdfkeywords={Physics-informed deep learning, Neural network training, Partial differential equations, Transportation models, Scalar conservation laws},
}

\begin{document}
\maketitle

\begin{abstract}
Since its introduction in 2017, physics-informed deep learning (PIDL) has garnered growing popularity in understanding the evolution of systems governed by physical laws in terms of partial differential equations (PDEs). However, empirical evidence points to the limitations of PIDL for learning certain types of PDEs. In this paper, we (a) present the challenges in training PIDL architecture, (b) contrast the performance of PIDL architecture in learning a first order scalar hyperbolic conservation law and its parabolic counterpart, (c) investigate the effect of training data sampling, which corresponds to various sensing scenarios in traffic networks, (d) comment on the implications of PIDL limitations for traffic flow estimation and prediction in practice. Detailed in the case study, we present the contradistinction in PIDL results between learning the traffic flow model (LWR PDE) and its variation with diffusion. The outcome indicates that PIDL experiences significant challenges in learning the hyperbolic LWR equation due to the non-smoothness of its solution. On the other hand, the architecture with parabolic PDE, augmented with the diffusion term, leads to the successful reassembly of the density data even with the shockwaves present.
\end{abstract}


\section{INTRODUCTION AND MOTIVATION} \label{sec:intr}

Traffic states such as vehicle velocity $v$, density $\rho$, and flow $f$ depict the condition of traffic operations on road infrastructures. An important task in transportation management is accurately capturing the current traffic state conditions to implement control measures such as ramp metering and tidal lanes during rush hours \cite{agarwal2015feedback, kong2009approach}. The practice of obtaining traffic state measurements is bounded by several constraints \cite{agarwal2015dynamic, xu2017real}. Firstly, sensing traffic state requires the installation of equipment along the roadway to detect the presence of vehicles. Costs of installation, calibration, and maintenance occur and limit this sensing application to a few selected locations in a complex road network. Secondly, inaccuracy and measurement noise is inherent in the process of acquiring traffic states through sensing devices. Considering the reliability of observed traffic data and handling missing values are common requirements to process traffic state measurement \cite{chen2001study}.

The observation data of traffic states can be harvested through another channel --- connected vehicles (CV) \cite{cardenas2016traffic}. In this approach, traffic conditions are perceived via the ubiquitous onboard sensors in connected vehicles zooming through traffic. CV has the potential of vastly broadening the detection area of traffic state on roadways \cite{uhlemann2015introducing}. However, this application scenario, in reality, is also restricted in several ways. Traffic states obtained through CV need to be broadcast and stored by roadside units or edge devices in a computing network \cite{du2017distributed}. The current penetration rates of both CV and the compatible communication system are insufficient to make it a reliable practice for procuring data on traffic states. Besides, the communication between CV and data infrastructure generates elevated demand for network resources in pursuit of robust on-site computing capability and low-latency data exchange \cite{si2016dave}. 

The estimation of traffic states, therefore, becomes an important practice for traffic planners and policy practitioners \cite{seo2017traffic}. It pertains to the inference of traffic state using data that is partially available, secured either through sensing devices or CV. A variety of traffic operations rely on this crucial assessment of traffic states \cite{greguric2020application, klepsch2017prediction}. Tasks as minute as determining the length of traffic cycles and as substantial as adding additional lanes on a highway all rely on the outcome of traffic state estimation (TSE). A variety of TSE methods exist in the literature and can be categorized into three groups: model-based, data-based, and streaming-data-based \cite{seo2017traffic}. Model-based approaches \cite{daganzo1994cell, munoz2006piecewise, payne1971model, aw2000resurrection} rely on a calibrated traffic flow model to estimate traffic states in unobserved areas. Data-based approaches \cite{li2013efficient, yin2012imputing, polson2017bayesian, azzouni2017long, wang2018deepstcl} utilize the insights derived from historical data and apply statistical or machine learning techniques to reconstruct and predict traffic states. The third category of streaming-data-based approaches \cite{bekiaris2016highway, qiu2010estimation} contains methods that do not necessitate a priori knowledge of historical observation, instead take advantage of real-time data, with a weak assumption of traffic relationship acquired from empirical evidence.

The restraints on obtaining traffic state data call for the adoption of TSE models that can exploit a limited amount of observed data for training and produce a precise estimation of traffic states. Among the \textcolor{black}{available approaches, deep learning (DL)} neural network is a powerful machine learning method increasingly used in many TSE applications \cite{xu2020ge, shi2021physics, mo2021physics}. However, DL neural network also comes with shortcomings, such as the high requirement of training data and computing power, \textcolor{black}{over-fitting, and transferability issues}, limiting its appeal for time-critical applications, which calls for the role of physics in aiding the training process of a neural network in TSE \cite{barreau2021physics, rempe2021estimating}.

\textbf{Physics-informed deep learning (PIDL)}, also referred to as physics-informed neural network (PINN), arms a neural network with the governing equations of a physical system \cite{raissi2017physics}. It empowers the deep learning neural network with knowledge of the underlying relationship in observed data to efficiently use the limited data input for estimation \textcolor{black}{and} prediction tasks \cite{raissi2017physics2}. Since the introduction of its architecture \cite{raissi2017physics2}, PIDL has been adopted in the field of traffic state estimation (TSE) \cite{huang2022physics, huang2020physics}. Researchers have experimented with PIDL to estimate both the traffic state and the fundamental diagram depicting the relationship between traffic states \cite{shi2021physics}. Second-order traffic models have also been considered in the application of PIDL for TSE \cite{shi2021physics2}. 

A diverse range of mechanical engineering and computational science applications have \textcolor{black}{also} been proposed, signifying its advantage in utilizing the governing equation to accurately \textcolor{black}{capture} the physical system. Among its diverse implementations in the mathematical domain, the PIDL approach has been adopted for solving the free boundary problems \cite{wang2021deep}, high-dimensional PDE \cite{han2018solving}, uncertainty quantification \cite{zhu2019physics}, \textcolor{black}{and} time-dependent stochastic PDE \cite{zhang2020learning}. In the field of fluid dynamics, PIDL is employed to model the velocity and pressure fields \cite{raissi2020hidden}, vortex-induced vibration \cite{raissi2019deep}, and fluid flows without the use of simulation data \cite{sun2020surrogate}. And on the engineering side, the modeling of cardiovascular flow \cite{kissas2020machine}, nano-optics \cite{chen2020physics}, and proxy modeling in solid mechanics \cite{haghighat2021physics} have all witnessed the effectiveness of the PIDL approach.

\textcolor{black}{
Variants of PIDL have been proposed to learn the solution of partial differential equations (PDEs). For instance, Galerkin method-based hp-VPINN was introduced to solve PDEs with non-smooth solutions \cite{kharazmi2021hp}. A Bayesian approach to PINN is presented for forward and inverse problems \cite{yang2021b}, and the idea of physics-informed adversarial training to solve PDE is proposed \cite{shekarpaz2022piat}. Particle swarm optimization is also put forward to PIDL training \cite{davi2022pso}.} 


Amid the PIDL applications that have been demonstrating encouraging results, the focal point is to use the neural network to learn the solutions of deterministic PDE commanding a physical system. As the PIDL model ciphers the underlying relationship between state variables, it incorporates the governing equation as a priori knowledge into calculating cost. If the PDE of interest is smooth and has a strong solution, at the same time, is paired with an adequate number of collocation points where the physics cost is optimized, then PIDL accordingly is capable of achieving good accuracy in learning the solution to the PDE. \textcolor{black}{However, recent research points to the limitations of PIDL for learning certain types of PDEs, such as hyperbolic conservation laws.}

\textbf{Scalar Conservation Laws in Traffic Flow Theory:} Lighthill-Whitham-Richards (LWR) is a one-dimensional scalar conservation law and a commonly-used transportation model. No strong solution exists to LWR, given it is a hyperbolic PDE \cite{agarwal2015feedback}. Certain classes of partial differential equations, \textcolor{black}{including} the LWR model, can be solved by the method of characteristics (details in Section~\ref{sec:potf}). In our previous work \cite{9294236, huang2022physics}, equipped with a realistic choice of data sampling on the interior traffic data (either Lagrangian data that come from connected vehicles (CV) or Eulerian data from roadside sensors and loop detectors), we have shown that PIDL can successfully and accurately learn the onerous task of reconstructing LWR PDE. However, no empirical evidence exists that PIDL can learn an LWR PDE with acceptable accuracy given only boundary and initial condition data, which is the default experimental setup for a variety of reconstruction problems \textcolor{black}{in transportation and other domains \cite{shi2021physics2, shi2021physics, raissi2017physics, raissi2017physics2}}. 


\textbf{Goals and Objectives:} The goal of this research is to understand the limitations of PIDL in traffic state estimation applications and initiate a dialogue on possible remedies in light of current literature. We aim to achieve the goal by addressing the following \textbf{\textit{research questions}}:

\begin{itemize}
     \item \textcolor{black}{Which form of the LWR model, hyperbolic PDE or parabolic PDE, is learned better by the PIDL algorithm?}
    \item \textcolor{black}{What are the factors behind the disparity in learning results?}
    \item \textcolor{black}{What is the effect of a diffusion term, which transforms the hyperbolic LWR PDE to a parabolic variation, in alleviating the weakness of PIDL?}
    \item \textcolor{black}{What are the inherent architectural issues in PIDL that inhibit it from working with hyperbolic conservation law?}
\end{itemize}


\textcolor{black}{
The research is divided into the following tasks: (a) exhibiting the contradistinction in terms of reconstruction accuracy between learning a commonly-used conservation law (LWR) in traffic flow theory, which is a first order hyperbolic PDE, and its second order parabolic counterpart by using a synthetic ring road density dataset, (b) further illustrating the limitations of PIDL in TSE with only the initial and boundary observations from the realistic field data (NGSIM), (c) understanding the impact of discontinuity in PDE solution, and the effect of a diffusion term on PIDL learning, (d) discussing the findings from the experimental results in light of existing literature and putting forward suggestions to improve PIDL results in TSE with the LWR PDE.
}


\textbf{Key features and contributions:}

\begin{enumerate}

  \item \textcolor{black}{We design a series of experiments by creating a circular testbed and by using realistic field (NGSIM) data to understand the performance of PIDL in TSE while incorporating the hyperbolic and parabolic versions of the physics.} 
    \item \textcolor{black}{We showcase the effect of the additional second order diffusion term in the LWR PDE on PIDL learning using only the initial and boundary observations from a ring road testbed.} 
    \item \textcolor{black}{We further investigate the contrast between PIDL learning the hyperbolic LWR PDE and its parabolic variant using a mix of Eulerian and Lagrangian training inputs (non-boundary data).}
    \item \textcolor{black}{Using the realistic NGSIM field data, we shed light on the limitations of PIDL in TSE and demonstrate that, on occasions, a numerical PDE solver outperforms PIDL.}
    \item \textcolor{black}{Finally, we discuss the reasoning behind the difficulties encountered when training a PIDL neural network with the LWR conservation law and list suggestions for improving the performance.}
\end{enumerate}


\paragraph*{Outline}
The rest of the manuscript is organized as follows: 
Section~\ref{sec:potf} gives the mathematical background of the scalar conservation laws, including the LWR model, strong and weak solutions, and \textcolor{black}{method of characteristics}. Section~\ref{sec:pidl} introduces the preliminaries of physics-informed deep learning (PIDL) for learning the conservation law-based traffic flow models. Section~\ref{sec:case} sets up a case study to reveal the challenges of PIDL with the first order hyperbolic LWR PDE. Section~\ref{sec:case2} further illustrates the limitations of PIDL with field data from the NGSIM dataset. \textcolor{black}{Section~\ref{sec:disc} discusses the implications from results and Section~\ref{sec:conc} concludes the paper with suggestions on future work.} 

\section{SCALAR CONSERVATION LAWS} \label{sec:potf}

In one dimension, a general form of the scalar conservation law is described by \eqref{eqn:scal_gene} where $f(u)$ is the flux function depending on the location $x$ and time $t$. The conserved variable is $u = u(x, \; t)$. Equation \eqref{eqn:scal_gene} becomes a Cauchy problem when the initial condition $u(x, \; 0) = u_0(x)$ is provided \cite{okutmucstur2019scalar}.

\begin{equation} \label{eqn:scal_gene}
    \partial_x f(u) + \partial_t u = 0, \;\;\; x \in \mathbb{R}, \; t \geq 0
\end{equation}

The flux function $f(u)$ can take many forms, the \textbf{linear advection equation} is a basic example of the scalar conservation law where the flux function $f(u)$ takes the form of $f(u) = \lambda u$ with a constant $\lambda$, shown in \eqref{eqn:scal_adve}.

\begin{equation} \label{eqn:scal_adve}
    \lambda \partial_x u + \partial_t u = 0, \;\;\; x \in \mathbb{R}, \; t \geq 0
\end{equation}

An initial value problem for the linear advection equation is given by \eqref{eqn:scal_adve_init} and its solution is exhibited in \eqref{eqn:scal_adve_solu}.

\begin{equation} \label{eqn:scal_adve_init}
    \begin{aligned}
        \partial_x f(u) + \partial_t u &= 0, \;\;\; x \in \mathbb{R}, \; t \geq 0 \\
        u(x, \; 0) = u_0(x) &= f(x_0), \;\;\; x \in \mathbb{R}
    \end{aligned}
\end{equation}

\begin{equation} \label{eqn:scal_adve_solu}
    u(x, \; t) = u_0(x - \lambda t), \;\;\;  t \geq 0
\end{equation}

A simple nonlinear particle differential equation is the \textbf{Burgers' equation}, and it is one of the commonly used models as the scalar conservation law. The classical form of Burgers' equation is presented in \eqref{eqn:scal_burg} where the $\epsilon \partial_{xx} u$ is the viscosity term. 

\begin{equation} \label{eqn:scal_burg}
    u \partial_x u + \partial_t u - \epsilon \partial_{xx} u = 0 , \;\;\; x \in \mathbb{R}, \; t \geq 0
\end{equation}

When $\epsilon = 0$, \eqref{eqn:scal_burg} becomes the \textbf{inviscid Burgers' equation}, and the flux function takes the form of $f(u) = u^2 / 2$, shown in \eqref{eqn:scal_burg_invi}. Plugging $\lambda = u$ into \eqref{eqn:scal_adve_solu} we get the solution of the inviscid Burgers' equation in \eqref{eqn:scal_burg_solu}.

\begin{equation} \label{eqn:scal_burg_invi}
    \partial_x (u^2 / 2) + \partial_t u = u \partial_x u + \partial_t u = 0 , \;\;\; x \in \mathbb{R}, \; t \geq 0
\end{equation}

\begin{equation} \label{eqn:scal_burg_solu}
    u(x, \; t) = u_0(x - ut), \;\;\;  t \geq 0
\end{equation}

The viscosity term is a diffusion term that flattens discontinuities and ensures a smooth solution.

\subsection{PHYSICS OF TRAFFIC FLOW}
Macroscopically, traffic flow is expressed using variables such as the mean velocity $v(x, \; t)$, a function of location $x$ and time $t$, traffic flow $q(x, \; t)$, and vehicle density $\rho(x, \; t)$. Spacing $s$ is the headway distance, which is the inverse of density $\rho$. Traffic flow is considered a continuum flow with a density profile associated with a compressible liquid \cite{lakhanpal2014numerical}. The traffic flow velocity is related to the density profile, time, and location.

\textbf{LWR conservation law} is a continuity equation, which holds for all macroscopic models, and formulates the conservation of traffic flow. This equation relates the change of density with the gradient of flow. When the flow is considered as a static function (portrayed by a fundamental diagram), it leads to a first order continuity equation, also referred to as Lighthill-Whitham-Richards (LWR) model \cite{lighthill1955kinematic} - which is a hyperbolic partial differential equation (PDE). Given a location $x$ between the start point $x_0$ and endpoint $x_n$, a timestamp $t$ between the initial time $t_0$ and the end time $t_m$, cumulative flow $N(x, \; t)$ depicts the number of vehicles that have reached location $x$ at time $t$. The flow $q(x, \; t)$ is the partial differential of $N(x, \; t)$ with respect to time $t$. Likewise, density $\rho(x, \; t)$ is the partial differential of $N(x, \; t)$ with respect to location $x$. 

The formulation of the LWR conservation law in the Eulerian coordinate system is shown in \eqref{eqn:lwr1}.  It is worth noting that data provided by connected vehicles (CV), such as vehicle velocity $v(n, \; t)$ and spacing $s(n, \; t)$ with respect to the vehicle identifier $n$ and time $t$, would be in Lagrangian coordinates \cite{contreras2015observability}. The Lagrangian formulation of the conservation law is shown in \eqref{eqn:lwr2}.

\vspace{1mm}
For $(x, \; t) \in [ \, x_0, \; x_n ] \, \times [ \, t_0, \; t_m ],   \;\;\;$
\begin{equation}  \label{eqn:lwr1}
    \frac{\partial q(x, \; t)}{\partial x} + \frac{\partial \rho(x, \; t)}{\partial t} = 0
\end{equation}

\vspace{1mm}
For $(n, \; t) \in [ \, n_0, \; n_i ] \, \times [ \, t_0, \; t_m ],   \;\;\;$ 
\begin{equation} \label{eqn:lwr2}
    \frac{\partial v(n, \; t)}{\partial n} + \frac{\partial s(n, \; t)}{\partial t} = 0
\end{equation}

\subsection{STRONG AND WEAK SOLUTIONS}

Given the initial value problem \eqref{eqn:scal_init}, 

\begin{equation} \label{eqn:scal_init}
    \begin{aligned}
        \partial_x f(u) + \partial_t u &= 0, \;\;\; x \in \mathbb{R}, \; t \geq 0 \\
        u(x, \; 0) &= u_0(x), \;\;\; x \in \mathbb{R}
    \end{aligned}
\end{equation}

If $u_0(x) \in C^1(\mathbb{R})$, then the initial condition $u_0(x)$ is continuously differentiable and \eqref{eqn:scal_init} becomes \eqref{eqn:scal_chai} by virtue of the chain rule.

\begin{equation} \label{eqn:scal_chai}
    \begin{aligned}
        f^\prime (u) \partial_x u + \partial_t u &= 0, \;\;\; x \in \mathbb{R}, \; t \geq 0 \\
        u(x, \; 0) &= u_0(x), \;\;\; x \in \mathbb{R}
    \end{aligned}
\end{equation}

In domain $\Omega \in \mathbb{R}$, a solution to \eqref{eqn:scal_chai}  is a \textbf{strong solution} (also referred to as ``classical solution'') if it satisfies \eqref{eqn:scal_init}, and is continuously differentiable \textcolor{black}{on} the domain $\Omega$.

When no strong solution to \eqref{eqn:scal_init} exists, the smoothness requirement can be relaxed to find \textbf{weak solutions}, even if these solutions are not differentiable or even continuous. Weak solutions \eqref{eqn:scal_weak} eliminate the derivative terms of $u$ and $f(u)$ to ease the smoothness requirement.

\textcolor{black}{Multiplying the scalar conservation law \eqref{eqn:scal_gene} with a function $\psi: \mathbb{R} \times \mathbb{R}^+ \to \mathbb{R}$, and given the initial condition $u(x, \; 0) = u_0(x)$, we have \eqref{eqn:scal_weak}.}

\begin{equation} \label{eqn:scal_weak}
    \int_{0}^{\infty} \int_{-\infty}^{\infty} (\partial_x f(u) + \partial_t u)\psi dxdt =
    \int_{0}^{\infty} \int_{-\infty}^{\infty} (f(u)\psi_x  + u \psi_t)dxdt + \int_{-\infty}^{\infty} u(x, \; 0)\psi(x)dx  = 0
\end{equation}

\textcolor{black}{
Notice in \eqref{eqn:scal_weak}, the requirement on smoothness is lessened as there are no derivative terms of $u$ and $f$. If \eqref{eqn:scal_weak} satisfies all $\psi(x)$, then $u(x, t)$ is the weak solution of \eqref{eqn:scal_init}.
}

It is worth noting that there is \textbf{no strong solution to the LWR conservation law}. Nevertheless, a diffusive term can be added to avoid breakdown and ensure a strong solution by making the hyperbolic conservation equation become a parabolic PDE. We will further discuss this in Section~\ref{sec:disc}. 

\subsection{METHOD OF CHARACTERISTICS}


The method of characteristics is used to solve quasilinear partial differential equations, converting the PDEs into ordinary differential equations (ODEs). Consider \eqref{eqn:scal_gene} and its solution $f(u) = u(x, \; t)$, let $x = x(t)$ solve the ODE in \eqref{eqn:scal_ode}:

\begin{equation} \label{eqn:scal_ode}
    \dot{x}(t) = u(x(t), \; t)
\end{equation}

From \eqref{eqn:scal_ode} \textcolor{black}{observe} that,

\begin{equation} \label{eqn:scal_ode_deri}
    \frac{d}{dt}u(x(t), \; t) = \textcolor{black}{\frac{dx}{dt}}u_x + u_t
\end{equation}

Combining \eqref{eqn:scal_gene} and \eqref{eqn:scal_ode_deri}, we reach \eqref{eqn:scal_ode_prop} that can propagate the solution \textcolor{black}{$u(x(t), \; t)$} with the initial condition $u_0(x)$.

\begin{equation} \label{eqn:scal_ode_prop}
    \frac{du}{dt} = 0, \; \frac{dx}{dt} = u
\end{equation}

A simple discontinuous solution of the conservation law \eqref{eqn:scal_init} is given by \eqref{eqn:scal_shoc}.

\begin{equation} \label{eqn:scal_shoc}
    u(x, \; t) = 
    \begin{cases}
        u_L, & x < \lambda t, \\
        u_R, & x \geq \lambda t, 
    \end{cases}
\end{equation}

If $u_L \neq u_R$, \eqref{eqn:scal_shoc} is termed a \textbf{shock wave}. \textcolor{black}{With the shock speed $\lambda$}, it connects $u_L$ to the value of $u_R$. Consider the following scalar Riemann problem \eqref{eqn:reim} where $\rho_L < \rho_R$:

\begin{equation} \label{eqn:reim}
\begin{aligned}
    v_f (1 - \frac{\rho(x, \; t)}{\rho_m}) \frac{\partial \rho(x, \; t)}{\partial x} + \frac{\partial \rho(x, \; t)}{\partial t}  = 0 \\
    \rho(x, \; 0) = 
    \begin{cases}
        \rho_L, & x < 0, \\
        \rho_R, & x \geq 0, 
    \end{cases}
\end{aligned}
\end{equation}

\begin{figure}[!htbp]
     \centering
     \begin{subfigure}[b]{0.45\textwidth}
         \centering
         \begin{tikzpicture}
            \draw[thick] (0.0, 0.0) -- (4.0, 0.0);
            \draw (0.0, 1.8) -- (0.4, 2.0);
            \draw (0.0, 1.6) -- (0.8, 2.0);
            \draw (0.0, 1.4) -- (1.2, 2.0);
            \draw (0.0, 1.2) -- (1.6, 2.0);
            \draw (0.0, 1.0) -- (2.0, 2.0);
            \draw (0.0, 0.8) -- (2.4, 2.0);
            \draw (0.0, 0.6) -- (2.8, 2.0);
            \draw (0.0, 0.4) -- (3.2, 2.0);
            \draw (0.0, 0.2) -- (3.6, 2.0);
            \draw (0.0, 0.0) -- (4.0, 2.0);
            \draw (0.4, 0.0) -- (3.6, 1.6);
            \draw (0.8, 0.0) -- (3.2, 1.2);
            \draw (1.2, 0.0) -- (2.8, 0.8);
            \draw (1.6, 0.0) -- (2.4, 0.4);
            \draw[thick, ->] (2.0, 0.0) -- (4.0, 2.0);
            \draw (2.2, 0.0) -- (2.4, 0.4);
            \draw (2.4, 0.0) -- (2.8, 0.8);
            \draw (2.6, 0.0) -- (3.2, 1.2);
            \draw (2.8, 0.0) -- (3.6, 1.6);
            \draw (3.0, 0.0) -- (4.0, 2.0);
            \draw (3.2, 0.0) -- (4.0, 1.6);
            \draw (3.4, 0.0) -- (4.0, 1.2);
            \draw (3.6, 0.0) -- (4.0, 0.8);
            \draw (3.8, 0.0) -- (4.0, 0.4);
        
        \end{tikzpicture}
         \caption{Shockwave solution}
         \label{fig:shock}
     \end{subfigure}
     \hfill
     \begin{subfigure}[b]{0.45\textwidth}
         \centering
         \begin{tikzpicture}
            \draw[thick] (0.0, 0.0) -- (4.0, 0.0);
            \draw (0.0, 1.6) -- (0.2, 2.0);
            \draw (0.0, 1.2) -- (0.4, 2.0);
            \draw (0.0, 0.8) -- (0.6, 2.0);
            \draw (0.0, 0.4) -- (0.8, 2.0);
            \draw (0.0, 0.0) -- (1.0, 2.0);
            \draw (0.2, 0.0) -- (1.2, 2.0);
            \draw (0.4, 0.0) -- (1.4, 2.0);
            \draw (0.6, 0.0) -- (1.6, 2.0);
            \draw (0.8, 0.0) -- (1.8, 2.0);
            \draw (1.0, 0.0) -- (2.0, 2.0);
            \draw (1.2, 0.0) -- (2.2, 2.0);
            \draw (1.2, 0.0) -- (2.4, 2.0);
            \draw (1.2, 0.0) -- (2.6, 2.0);
            \draw (1.2, 0.0) -- (2.8, 2.0);
            \draw (1.2, 0.0) -- (3.0, 2.0);
            \draw (1.2, 0.0) -- (3.2, 2.0);
            \draw (1.4, 0.0) -- (3.4, 2.0);
            \draw (1.6, 0.0) -- (3.6, 2.0);
            \draw (1.8, 0.0) -- (3.8, 2.0);
            \draw (2.0, 0.0) -- (4.0, 2.0);
            \draw (2.2, 0.0) -- (4.0, 1.8);
            \draw (2.4, 0.0) -- (4.0, 1.6);
            \draw (2.6, 0.0) -- (4.0, 1.4);
            \draw (2.8, 0.0) -- (4.0, 1.2);
            \draw (3.0, 0.0) -- (4.0, 1.0);
            \draw (3.2, 0.0) -- (4.0, 0.8);
            \draw (3.4, 0.0) -- (4.0, 0.6);
            \draw (3.6, 0.0) -- (4.0, 0.4);
            \draw (3.8, 0.0) -- (4.0, 0.2);
        \end{tikzpicture}
         \caption{Rarefaction solution}
         \label{fig:rare}
     \end{subfigure}
    \caption{Ring Road Vehicle Density Experimental Data}
    \label{fig:solution}
\end{figure}
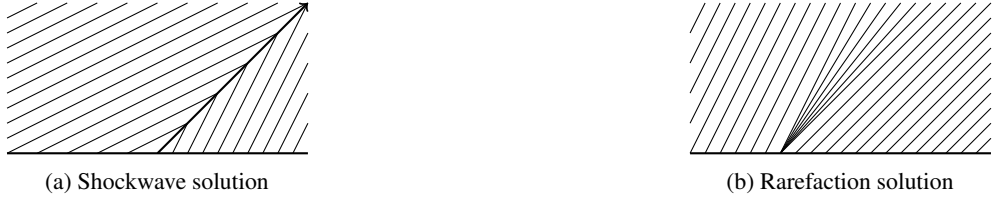

The characteristic speed at $t = 0$ is given in \eqref{eqn:speed}. As $\rho_L < \rho_R$, the characteristic speed $\lambda(\rho_L)$ on the left is greater than the right $\lambda(\rho_R)$ and develops a shock curve, shown in Fig.~\ref{fig:shock}.

\begin{equation} \label{eqn:speed}
    \lambda(\rho) = f^\prime(\rho) = v_f (1 - 2 \frac{\rho}{\rho_m})
\end{equation}

If we modify the problem in \eqref{eqn:reim} with the initial condition that $\rho_L > \rho_R$, the value of the characteristic speed on the left $\lambda(\rho_L)$ will be smaller than the value of speed on the right $\lambda(\rho_L)$. One of the possible solutions to \eqref{eqn:reim} with $\rho_L > \rho_R$ is the symmetry rarefaction solution, which is stable to perturbation to the initial data. The solution is given in \eqref{eqn:rare}, and shown in Fig.~\ref{fig:rare}.

\begin{equation} \label{eqn:rare}
\begin{aligned}
\textcolor{black}{
    \rho(x, \; t) = 
    \begin{cases}
        \rho_L, & \frac{x}{t} < \lambda(\rho_L), \\
        \frac{\lambda(\rho_R) - \lambda(\rho_L)}{\rho_R - \rho_L} \frac{x}{t}, &  \lambda(\rho_L) \leq  \frac{x}{t} < \lambda(\rho_R),\\
        \rho_R, &  \lambda(\rho_R) \leq \frac{x}{t}\\
    \end{cases}
}
\end{aligned}
\end{equation}

\section{PHYSICS-INFORMED DEEP LEARNING} \label{sec:pidl}

Development in deep learning (DL) neural networks has made it a suitable tool in the computational modeling of a physical system, which is often governed by complex non-linear functions \cite{sirignano2018dgm}. When mean square error (MSE) is used as the measurement of cost in a DL neural network, the cost function can be formulated as \eqref{eqn:cos_dl}, in which $N$ is the number of outputs, and $\hat{\mathbf{u}}(x, \ y, \ z, \ t)$ is the prediction of the variable ${\mathbf{u}}(x, \ y, \ z, \ t)$.

\begin{align} \label{eqn:cos_dl}
    J &= MSE_{(\hat{\mathbf{u}}(x, \ y, \ z, \ t), \; \mathbf{u}(x, \ y, \ z, \ t))} \nonumber \\[2pt]
    &= \frac{1}{N} \sum_{k=1}^{N}\left|\mathbf{u}(x, \ y, \ z, \ t) - \hat{\mathbf{u}}(x, \ y, \ z, \ t)\right|^2 
\end{align}

Automatic differentiation (AD) computes a state variable's partial derivatives with respect to its spatial and temporal independent variables \cite{baydin2017automatic}. Through the layers of a neural network, an output $\mathbf{u}(x, \ y, \ z, \ t)$ can be presented as a nested function of input variables $x, y, z$, and $t$. By applying the chain rule, automatic differentiation yields accurate derivatives of the training cost with respect to the parameters of the neural network \cite{rao2020physics}.

To evaluate the outputs from the neural network in terms of compliance with the governing physical laws, the physics cost is computed at a set of spatiotemporal points $(x, \ y, \ z, \ t)$, termed as collocation points, which can be chosen by Latin hypercube sampling (LHS) \cite{mckay2000comparison}. To differentiate the deep learning cost (DL-cost) in \eqref{eqn:cos_dl} and the physics-cost, we use $J_{DL}$ to depict the DL-cost, and $J_{PHY}$ as the penalty of incompliance to the physics. $J_{DL}$ and $J_{PHY}$ are computed individually in \eqref{eqn:cos_pidl}. $N_u$ symbolizes the number of training samples, and $N_f$ is the number of collocation points.

\begin{equation} \label{eqn:cos_pidl}
   \left\{ \,
       \begin{IEEEeqnarraybox}
           [\IEEEeqnarraystrutmode
           \IEEEeqnarraystrutsizeadd{7pt}{7pt}][c]{lr}
           J_{DL} = \frac{1}{N_u} \sum_{k=1}^{N_u}\left|\mathbf{u}(x, \ y, \ z, \ t) - \hat{\mathbf{u}}(x, \ y, \ z, \ t)\right|^2 
           \\[2pt]
           J_{PHY} = \frac{1}{N_f} \sum_{k=1}^{N_f}\left|\mathbf{f}(x, \ y, \ z, \ t) \right|^2 
       \end{IEEEeqnarraybox}
   \right.
\end{equation}

\begin{figure*}[htbp]
    \centerline{\includegraphics[width=0.9\textwidth]{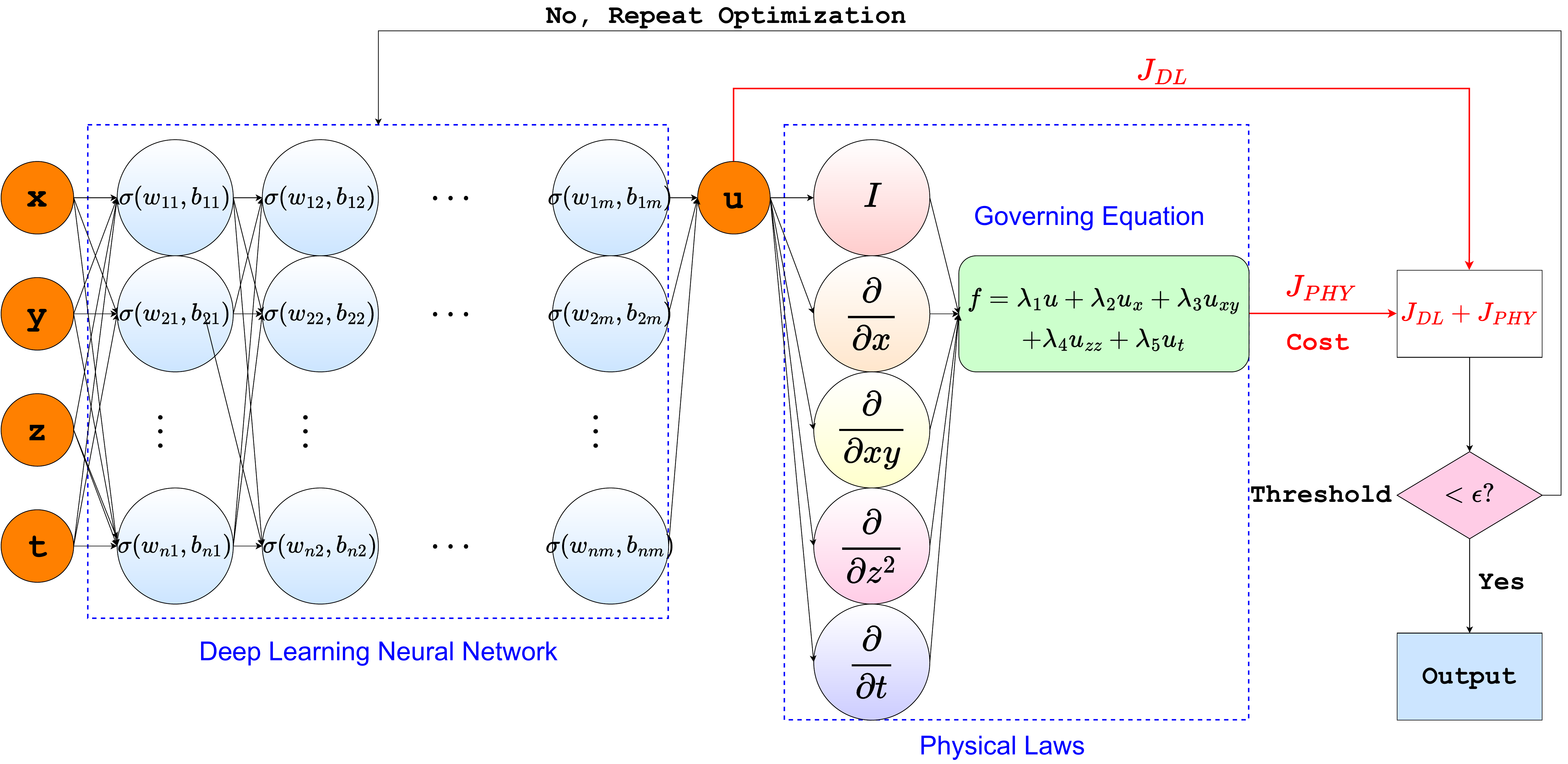}}
    \caption{Architecture of a Physics-informed Deep Learning (PIDL) Neural Network}
    \label{fig:pinn}
\end{figure*}

In \eqref{eqn:cos_pidl}, cost function $\mathbf{f}(x, \ y, \ z, \ t)$ is configured to quantify the non-compliance of the physics. As in the exemplary Fig.~\ref{fig:pinn}, which outlines the architecture of a PIDL neural network, the cost function of $J_{PHY}$ is given as $\mathbf{f} = \lambda_1u + \lambda_2u_x + \lambda_3u_{xy} + \lambda_4u_{zz} + \lambda_5u_{t}$. The cost terms $u_x, u_{xy}, u_{zz},$ and $u_{t}$ are partial derivatives of the output $\hat{\mathbf{u}}(x, \ y, \ z, \ t)$ with respect to a few combinations of input coordinates $(x, \ y, \ z, \ t)$. The $\lambda$ parameters represent the weights of cost terms. 

\subsection{PIDL FOR TRAFFIC STATE ESTIMATION}

This section provides details of the PIDL approach for traffic state estimation (TSE) by using LWR PDE and Greenshield's fundamental diagram. Other physical models, such as \textcolor{black}{second order} flow models or discretized first order models can be obtained by following the steps laid out in this section. PIDL empowers a DL neural network with the system's governing physical laws as priori knowledge \cite{raissi2017physics}. The fundamental diagram of traffic flow and the conservation law serve as meaningful know-how in training a neural network to recognize the underlying relationship between traffic variables. 

Several fundamental diagrams exist and are used accordingly depending on the situation. The commonly-used ones are: piece-wise affine speed-density relationship \cite{munoz2006piecewise}, triangular fundamental diagram \cite{whitham1974linear}, etc. Greenshield's fundamental diagram \cite{greenshields1935study}, is one of the most utilized and simplest models in traffic flow theory. It makes an untangling assumption that the mean velocity has a linear relationship with the density. The relationship between traffic variables is described in \eqref{eqn:gree}, where $\rho_m$ is the maximum density, $s_m$ is the minimum spacing, and $v_f$ is the free-flow speed. 

\begin{equation} \label{eqn:gree}
    \left\{ \,
        \begin{IEEEeqnarraybox}
            [\IEEEeqnarraystrutmode
            \IEEEeqnarraystrutsizeadd{7pt}{7pt}][c]{rCl}      
            q(x, \; t) & = & \rho(x, \; t) \; v_f \left(1 - \frac{\rho(x, \; t)}{\rho_m}\right) \\[2pt]
            v(x, \; t)  & = & v_f \left(1 - \frac{\rho(x, \; t)}{\rho_m} \right)
        \end{IEEEeqnarraybox}
    \right.
\end{equation}

Plugging the relationship between variables $\rho$, $v$, and $q$ from \eqref{eqn:gree}  into the Eulerian formulation of conservation law in \eqref{eqn:lwr1}, the physical law can be written as \eqref{eqn:lwrv} and \eqref{eqn:lwrd}.

\vspace{1mm}
For $(x, \; t) \in [ \, x_0, \; x_n ] \, \times [ \, t_0, \; t_m ],   \;\;\;$
\begin{equation} \label{eqn:lwrv}
    \rho_m \left(1 - \frac{2v(x, \; t)}{v_f}\right)  \frac{\partial v(x, \; t)}{\partial x}   - \frac{\rho_m}{v_f} \frac{\partial v(x, \; t)}{\partial t} = 0
\end{equation}

\vspace{1mm}
For $(x, \; t) \in [ \, x_0, \; x_n ] \, \times [ \, t_0, \; t_m ],   \;\;\;$
\begin{equation} \label{eqn:lwrd}
    v_f \left(1 - \frac{2\rho(x, \; t)}{\rho_m}\right)  \frac{\partial \rho(x, \; t)}{\partial x} + \frac{\partial \rho(x, \; t)}{\partial t} = 0
\end{equation}

\noindent Observe that both the equations provide the same physical law - the only difference is their dependent variable. Equation \eqref{eqn:lwrv} formulates the law in terms of velocity $v(x, \; t)$, whereas \eqref{eqn:lwrd} formulates it in terms of density $\rho(x, \; t)$.

\textcolor{black}{
Both \eqref{eqn:lwrv} and \eqref{eqn:lwrd} are hyperbolic PDEs. A \textcolor{black}{second order} diffusive term can be added to make the PDE become parabolic and secure a strong solution. For example, \eqref{eqn:lwrd} will become \eqref{eqn:lwrd_para} where $\epsilon$ is a constant of a small value.
}

\vspace{1mm}
\textcolor{black}{
For $(x, \; t) \in [ \, x_0, \; x_n ] \, \times [ \, t_0, \; t_m ],   \;\;\;$
\begin{equation} \label{eqn:lwrd_para}
    v_f \left(1 - \frac{2\rho(x, \; t)}{\rho_m}\right)  \frac{\partial \rho(x, \; t)}{\partial x} + \frac{\partial \rho(x, \; t)}{\partial t} = \epsilon \frac{\partial^2 \rho(x, \; t)}{\partial x^2}
\end{equation}
}

\textcolor{black}{
The \textcolor{black}{second order} diffusion term ensures the solution of PDE is continuous and differentiable, avoiding the breakdown and discontinuity in the solution to the PDE.
}

\subsection{TRAINING DATA \textcolor{black}{AND COST FUNCTIONS}}

The cost function of a PIDL neural network reconstructing a density-field $\rho(x \; t)$ is written as \eqref{eqn:cosd}, where $J_{DL}$ is computed at observation points $\mathcal{O} = \{(x_{o}^{j}, t_{o}^{j}) | j = 1, 2, \cdots, N_{o}\}$, and $J_{PHY}$ is obtained at the collocation points $\mathcal{C} = \{(x_{c}^{j}, t_{c}^{j}) | j = 1, 2, \cdots, N_{c}\}$. 

\begin{equation} \label{eqn:cosd} 
   \left\{ \,
       \begin{IEEEeqnarraybox}
           [\IEEEeqnarraystrutmode
           \IEEEeqnarraystrutsizeadd{7pt}{7pt}][c]{lr}
           J_{DL} = \frac{1}{N_{o}}\sum_{j=1}^{N_{o}}\left|\rho(x_{o}^{j}, t_{o}^{j}) - \hat{\rho}(x_{o}^{j}, t_{o}^{j})\right|^2 \\[2pt]
           J_{PHY} = \frac{1}{N_{c}}\sum_{j=1}^{N_{c}}\Big|v_f (1 - \frac{2\hat{\rho}(x_{c}^{j}, t_{c}^{j})}{\rho_m})\frac{\partial \hat{\rho}(x_{c}^{j}, t_{c}^{j})}{\partial x} \\
           \hspace{2.6cm} + \frac{\partial \hat{\rho}(x_{c}^{j}, t_{c}^{j})}{\partial t} \Big|^2 
       \end{IEEEeqnarraybox}
   \right.
\end{equation}


Weights can be assigned to the cost terms of $J_{DL}$ and $J_{PHY}$ in constituting the cost function of PIDL, as composed in \eqref{eqn:cos_weig}.

\begin{equation} \label{eqn:cos_weig}
    J = \mu_1 * J_{DL} + \mu_2 * J_{PHY}
\end{equation}

\noindent where weight parameters $\mu_1$ and $\mu_2$ adjust the scales of the DL-cost and the physics-cost.


\subsection{TRAINING APPROACHES} 

\textcolor{black}{
There are many optimization algorithms to train a neural network, here we provide brief information on the following training approaches we used in this work. 
}

\subsubsection{L-BFGS-B}
Limited memory, boundary constraints Broyden–Fletcher–Goldfarb–Shanno algorithm \cite{liu1989limited} is one of the default optimizers of \texttt{scipy.optimize.minimize} in the scientific computing library SciPy \cite{virtanen2020scipy}. Under the default setting, the optimization process terminates when the difference of cost $ftol$ between iterations is less than 2.22e-16.

\subsubsection{Adam}
Adaptive moment estimation (Adam) \cite{kingma2014adam} takes the advantages in the Momentum \cite{qian1999momentum} and the RMSProp \cite{tieleman2012lecture} optimization algorithms by monitoring the accumulation of both the gradient and the squared gradient, using $\beta_1$ and $\beta_2$ as the decay terms shown in \eqref{eqn:adam}. 


\begin{align} \label{eqn:adam}
    \Delta \theta_{i, t} &= \nabla \mathit{J}(\theta_{i, t}) \nonumber \\[2pt]
    \mathit{G}_{i, t} &= \beta_1 \mathit{G}_{i, t-1} + (1 - \beta_1) \Delta \theta_{i, t} \nonumber \\[2pt]
    \mathit{E}_{i, t} &= \beta_2 \mathit{E}_{i, t-1} + (1 - \beta_2) (\Delta \theta_{i, t})^2 \nonumber \\[2pt]
    \theta_{i, t+1} &= \theta_{i, t} - \alpha \frac{\mathit{E}_{i, t}}{\mathit{G}_{i, t} + \epsilon} \Delta \theta_{i, t}
\end{align}

\subsection{ERROR METRIC} \label{sssec:difu}

\textcolor{black}{
\textbf{\textit{Relative $\mathcal{L}_2$ Error}} - To evaluate the inaccuracy of a trained neural network, we establish the relative $\mathcal{L}_2$ error term, which is defined in \eqref{eqn:rl2e}, using vehicle density $\rho(x, \; t)$ as the exemplary dependent variable.
}

\textcolor{black}{
\begin{align} \label{eqn:rl2e}
    \mathcal{L}_2^{error}
    &=  \frac{\norm{\mathbf{P} - \mathbf{\hat{P}}}_F}{\norm{\mathbf{P}}_F} 
    \nonumber \\[2pt]
    &= \frac{\sqrt{\sum_{j=1}^{N_1 \cdot N_2} \left|\hat{\rho}(x^{(j)}, t^{(j)}) - \rho(x^{(j)}, t^{(j)})\right|^2}}{\sqrt{\sum_{j=1}^{N_1 \cdot N_2}\left|\rho(x^{(j)}, t^{(j)})\right|^2}} 
\end{align}
}

\textcolor{black}{
\noindent where $\mathbf{P}$ is the matrix form of $\rho(x, \; t)$, and $\mathbf{\hat{P}}$ is the neural network's estimation of $\mathbf{P}$. After discretizing the density field $\rho(x, \; t)$ in time and space, $N_1$ is the number of temporal bins and $N_2$ is the number of spatial bins. 
}

\section{CASE STUDY \textcolor{black}{I - INSIGHTS FROM TESTBED}} \label{sec:case}

\begin{figure}[htbp]
    \centerline{\includegraphics[width=0.25\textwidth]{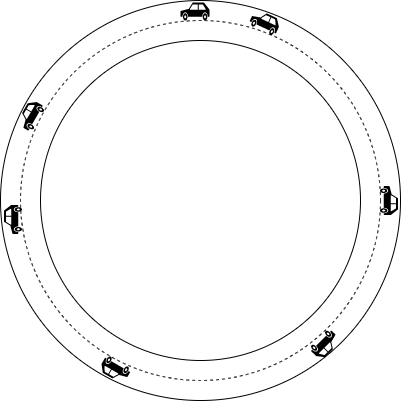}}
    \caption{Ring Road Configuration of the Datasets}
    \label{fig:ring_road}
\end{figure}

In this case study, we compare the PIDL reconstruction accuracy between learning the LWR conservation law (hyperbolic) and its parabolic form. The datasets used in this case study are synthetic vehicle density datasets generated on a ring road (represented by Fig.~\ref{fig:ring_road}). We configure all the neural networks with the same learning architecture (equal numbers of layers, same number of neurons on each layer, etc.). 10000 collocation points are assigned in the density field to compute the physics-cost. The learning rate of Adam is set to 0.001, and the number of training iterations is set to 8000. 

\subsection{DATASET}

The datasets for this case study simulate vehicular traffic on a ring road. The location $x$ and time $t$ are normalized as $x, \; t \in [0, \, 1.0] \, \times [0, \, 3.0]$. The road is evenly divided into 240 spatial units with $\Delta x = 1 / 240$, and time is similarly separated into 960 units and each timestep represents the progression of $\Delta t = 1 / 320$.

\begin{figure}[!htbp]
     \centering
     \begin{subfigure}[b]{0.45\textwidth}
         \centering
         \includegraphics[width=\textwidth]{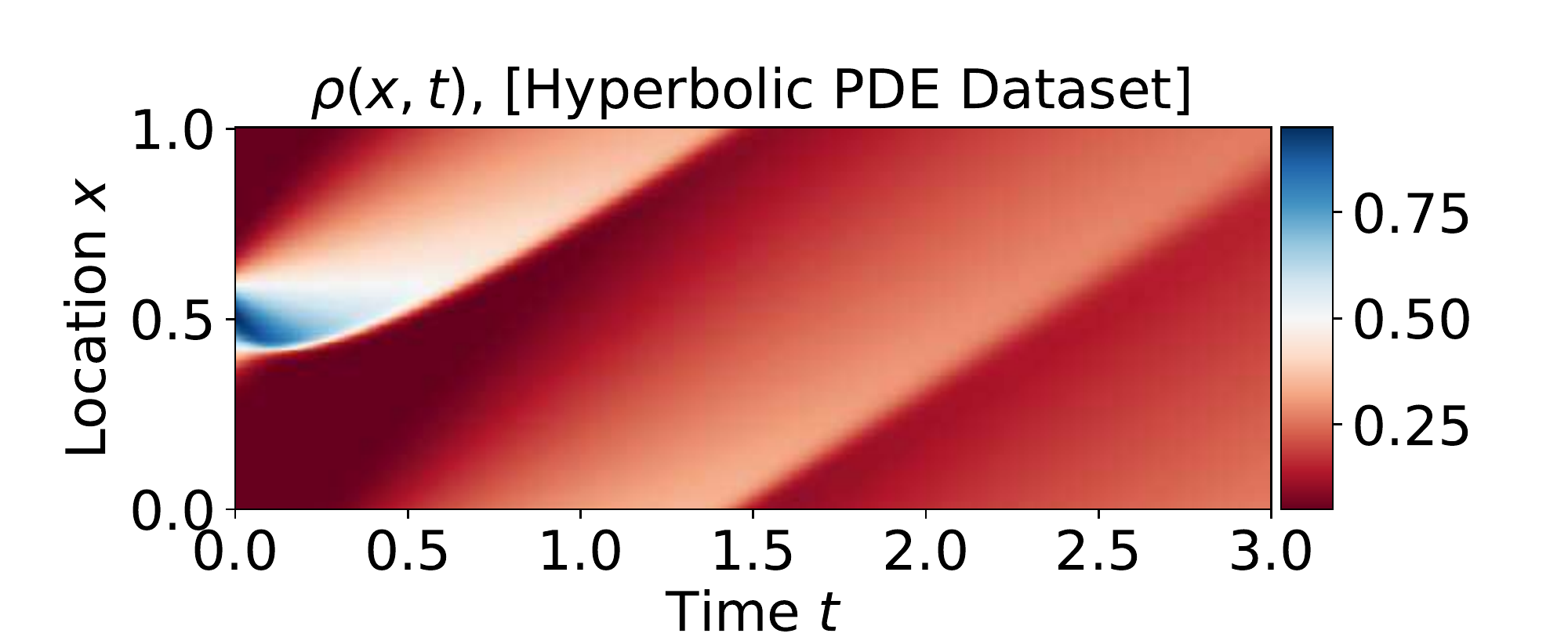}
         \caption{Dataset with \textbf{hyperbolic} conservation law}
         \label{fig:data_ring1}
     \end{subfigure}
     \hfill
     \begin{subfigure}[b]{0.45\textwidth}
         \centering
         \includegraphics[width=\textwidth]{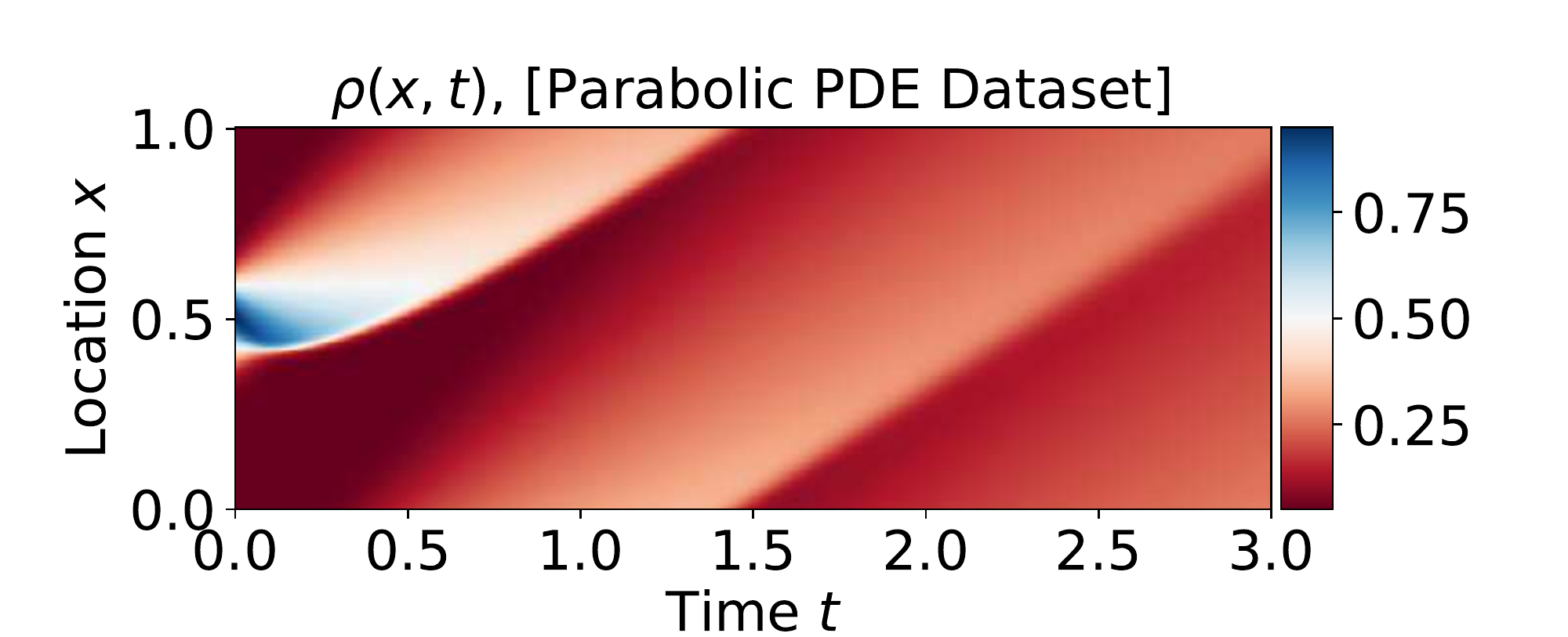}
         \caption{Dataset with \textbf{parabolic} conservation law}
         \label{fig:data_ring2}
     \end{subfigure}
    \caption{Ring Road Vehicle Density Experimental Data}
    \label{fig:data_ring}
\end{figure}

The traffic state variable of interest in this case study is the vehicle density $\rho(x, \; t)$. The assumed physical model is LWR conservation law, paired with Greenshield's fundamental diagram (FD) formulated in \eqref{eqn:gree}. The values of the free flow speed $v_f$ and the maximum density $\rho_m$ are normalized as well as both are set to $1$. We first generate the dataset shown in Fig.~\ref{fig:data_ring1}, by using the LWR conservation law (hyperbolic PDE) in \eqref{eqn:lwrd}. Based on the same initial and boundary conditions, and only adding a diffusion term to the LWR PDE to make it parabolic, as explained in \eqref{eqn:lwrd_para}, we are able to configure the dataset illustrated in Fig.~\ref{fig:data_ring2} by the parabolic form of LWR PDE. The \textcolor{black}{relative mean squared value of the difference between these two datasets in Fig.~\ref{fig:data_ring} is $0.35\%$, indicating almost identical density values of the datasets.}

\subsection{SELECTION OF LEARNING DATA INSTANCES}
\begin{figure}[htbp]
     \centering
     \begin{subfigure}[b]{0.45\textwidth}
         \centering
         \includegraphics[width=\textwidth]{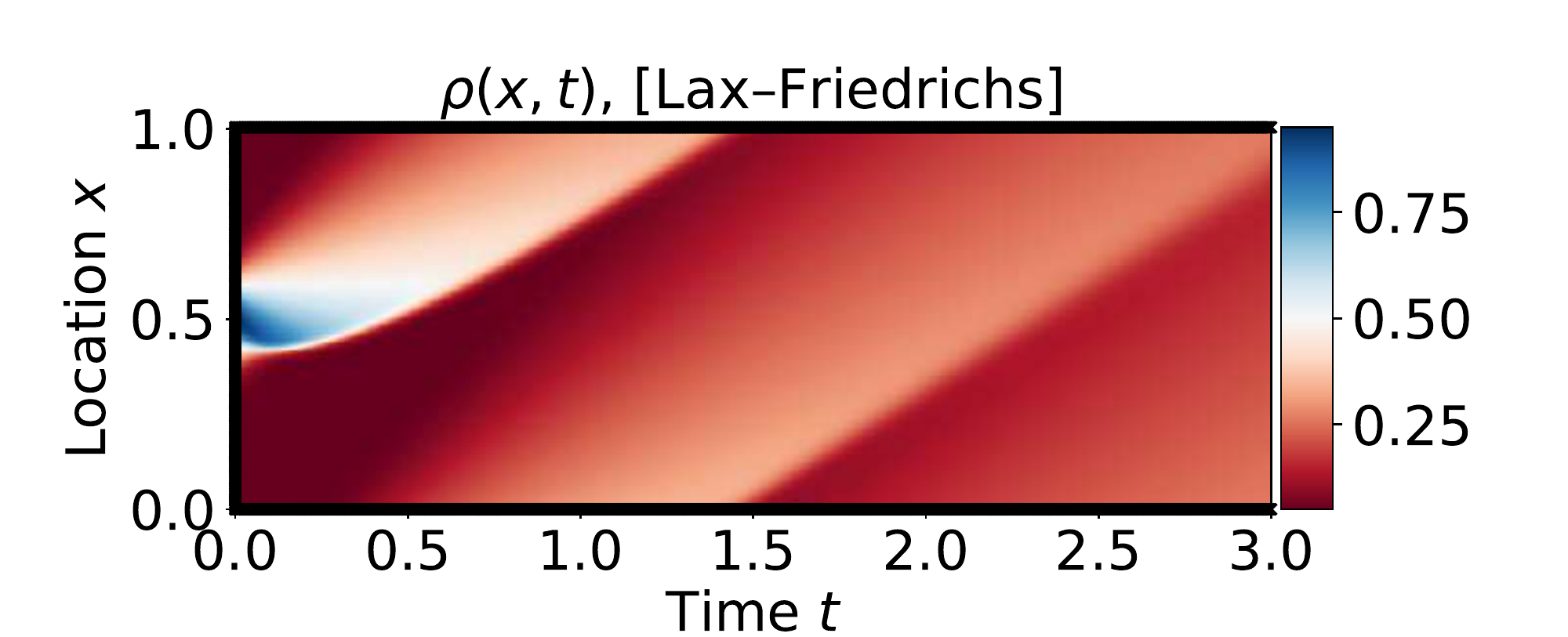}
         \caption{Lax–Friedrichs', All Boundary \& Initial Observations}
         \label{fig:data_repr1}
     \end{subfigure}
     \hfill
     \begin{subfigure}[b]{0.45\textwidth}
         \centering
         \includegraphics[width=\textwidth]{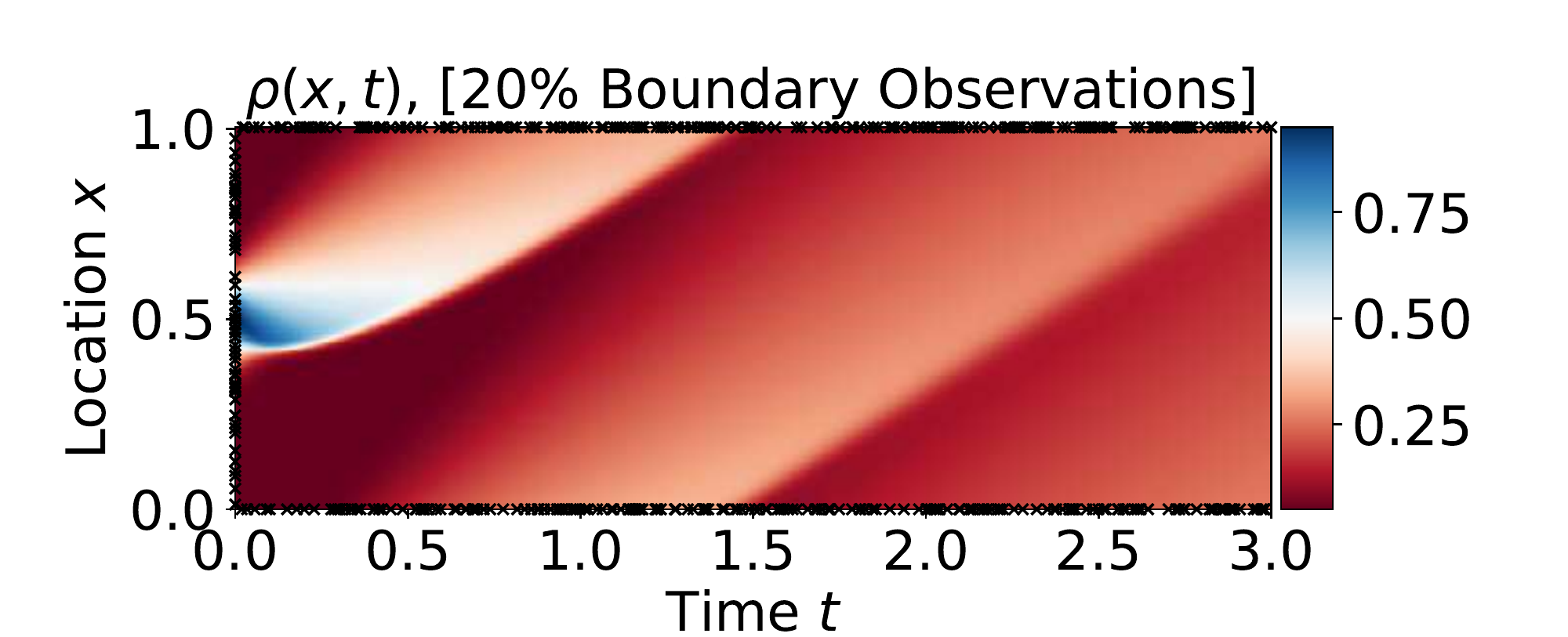}
         \caption{PIDL, 20\% Boundary \& Initial Observations}
         \label{fig:data_repr2}
     \end{subfigure}
     \hfill
     \begin{subfigure}[b]{0.45\textwidth}
         \centering
         \includegraphics[width=\textwidth]{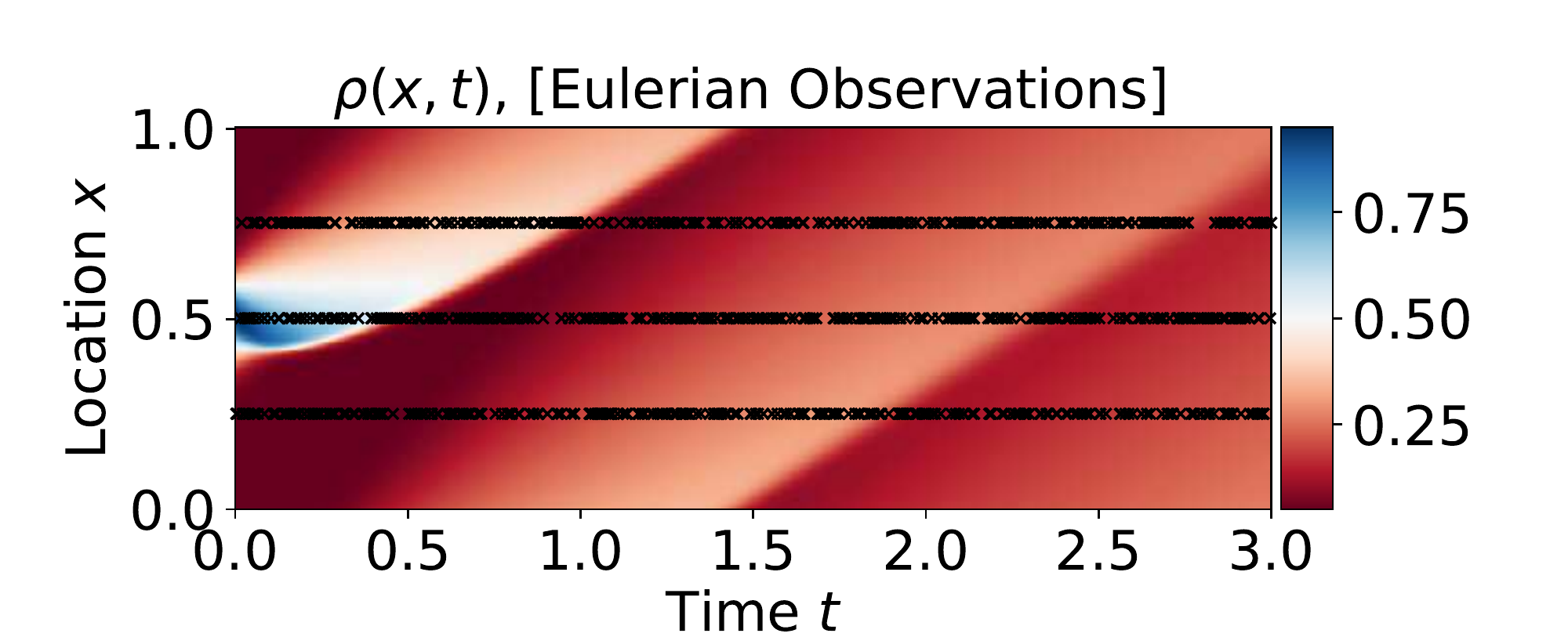}
         \caption{Eulerian Observations}
         \label{fig:data_repr3}
     \end{subfigure}
     \hfill
     \begin{subfigure}[b]{0.45\textwidth}
         \centering
         \includegraphics[width=\textwidth]{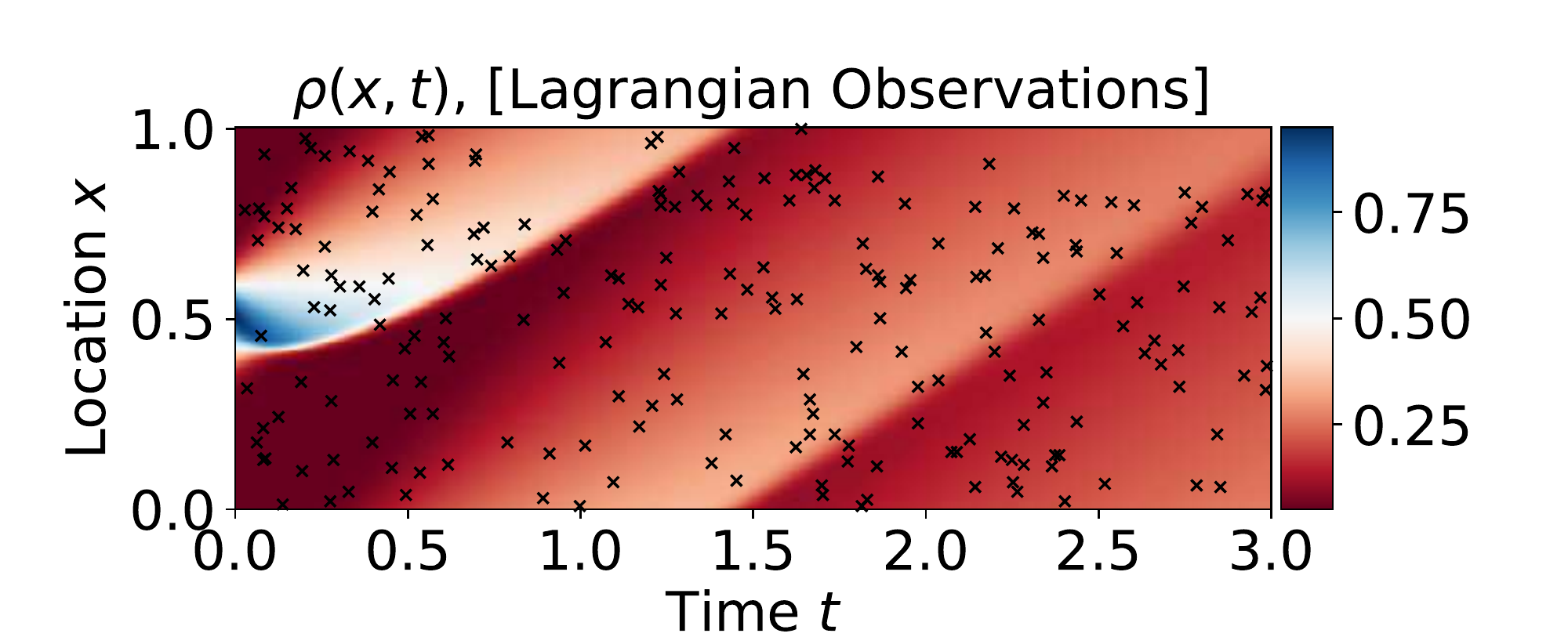}
         \caption{Lagrangian Observations}
         \label{fig:data_repr4}
     \end{subfigure}
        \caption{Selection of Learning Data Instances}
        \label{fig:data_repr}
\end{figure}

\textcolor{black}{
The sampling of available data for learning is an important aspect of machine learning. Data collection and sensing may introduce potential biases, often due to human factors and sensing limitations. These biases can persist in the models that are trained on the data, highlighting the importance of selecting appropriate subsets of the data to minimize these biases. However, in many instances, the selection of training instances is limited by the availability and positions of the sensors. E.g. for traffic sensing, the sensors are either fixed on the roadside at fixed intervals (e.g., loop detectors) or they are moving with the traffic stream (probe vehicles, CVs).}
Fig.~\ref{fig:data_repr} demonstrates the sampling \textcolor{black}{cases} of traffic state data. \textcolor{black}{Numerical PDE solvers such as} Lax-Friedrichs' numerical scheme \cite{shampine2005two} \textcolor{black}{can be used as} a state reconstruction \textcolor{black}{tool; however, it requires} the complete information on the initial and boundary \textcolor{black}{conditions as shown in Fig.~\ref{fig:data_repr1} which is not practically feasible. On the other hand, the PIDL approach can utilize any given amount of inputs from the boundaries for training;} in Fig.~\ref{fig:data_repr2}, 20\% of the initial and boundary data are shown as an exemplar training setting. Fig.~\ref{fig:data_repr3} represents the Eulerian traffic data that can be gathered from roadside sensors or loop detectors installed at predetermined locations along the road infrastructure (shown at $x = 0, 0.25, 0.5, 0.75, 1.0$ in this instance). Finally, Fig.~\ref{fig:data_repr4} exhibits the Lagrangian traffic data that can be collected by connected vehicles, which can muster traffic state information at various locations along the vehicle trajectories. Notice the gaps between the Eulerian data in Fig.~\ref{fig:data_repr3}; these represent the occasional sensor failures or malfunctions, resulting in data loss in this scenario.  

Given the sampling choices, we designate two subsets of training data inputs about the vehicle density $\rho (x, t)$: (1) initial and boundary condition data inputs and (2) interior data inputs. The interior data of the density field can be collected by either roadside detectors (Eulerian) or connected vehicles (Lagrangian). In the case study, we select the Lagrangian data from CV as the interior data.

\begin{enumerate}
    \item Initial condition data are the vehicle density values $\rho(x, \; 0)$ as $t = 0, x \in [0, 1.0]$. Boundary condition data include vehicle density values at the first location $\rho(0, \; t), \; t \in [0, 3.0]$, and at the last location $\rho(1.0, \; t), \; t \in [0, 3.0]$. 
    \item Interior data (CV data in the case study) $\rho (x, t)$ comes from the CV fleet in the traffic. They can be gathered at any randomly selected location along the vehicle trajectory and reflect the density value $\rho(x, \; t)$, given $x \in [0, \, 1.0]$ and $t \in [0, \, 3.0] \,$.
\end{enumerate}

The initial condition data on vehicle density can be registered through a still image, recorded by devices such as roadside video cameras or drones. The boundary condition data can be obtained from a stationary detector deployed along a freeway at the start location $x = 0$, and the end location $x = 1.0$ (normalized).

\subsection{RECONSTRUCTION WITH INITIAL AND BOUNDARY INPUTS}

\begin{figure*}[htbp]
     \centering
     \begin{subfigure}[b]{0.45\textwidth}
         \centering
         \includegraphics[width=3.3in]{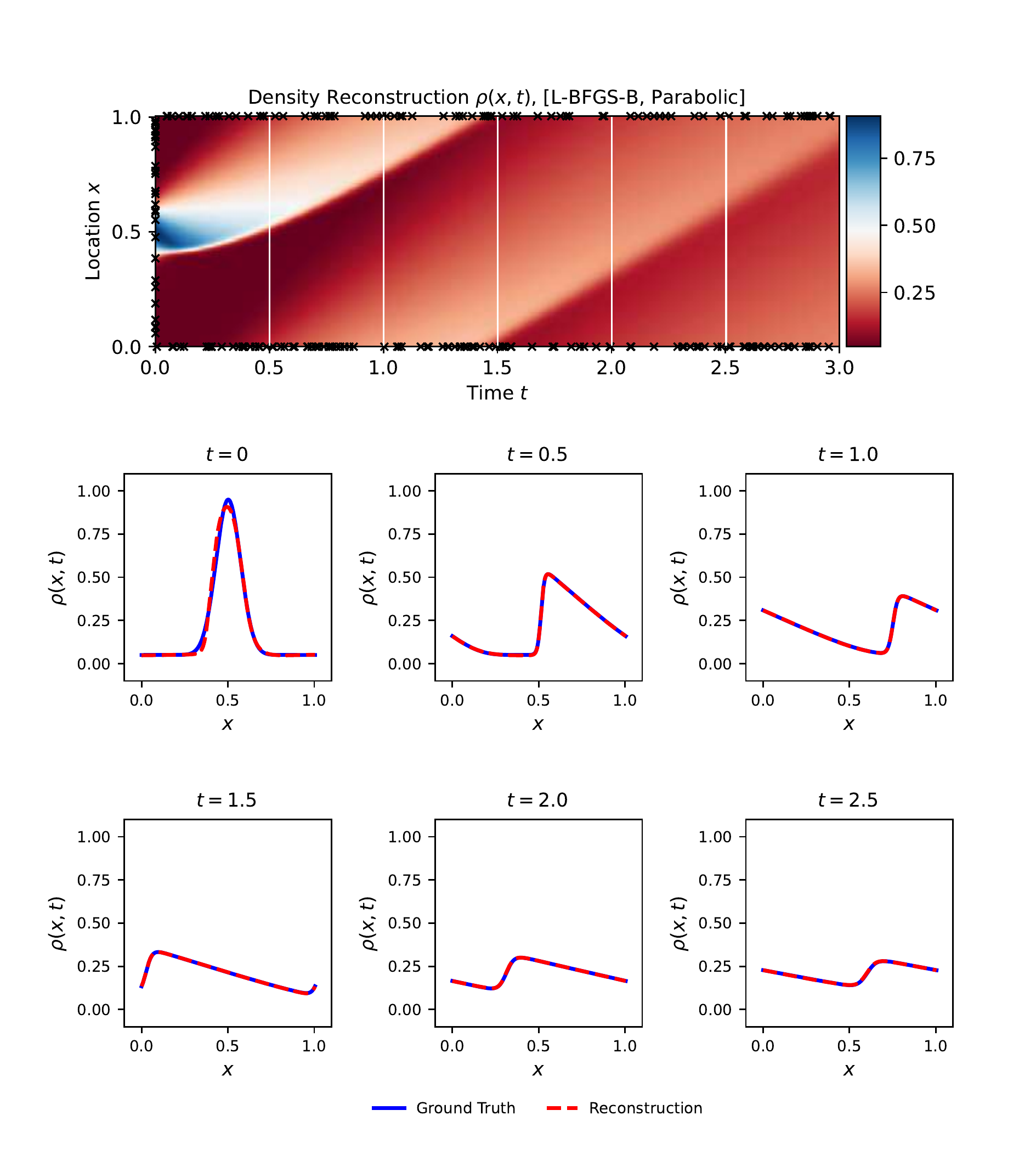}
         \caption{Learning \textbf{parabolic conservation law} using L-BFGS-B optimizer, relative $\mathcal{L}_2$ error: $0.0164$}
         \label{fig:reco_bund1}
     \end{subfigure}
     \hfill
     \begin{subfigure}[b]{0.45\textwidth}
         \centering
         \includegraphics[width=3.3in]{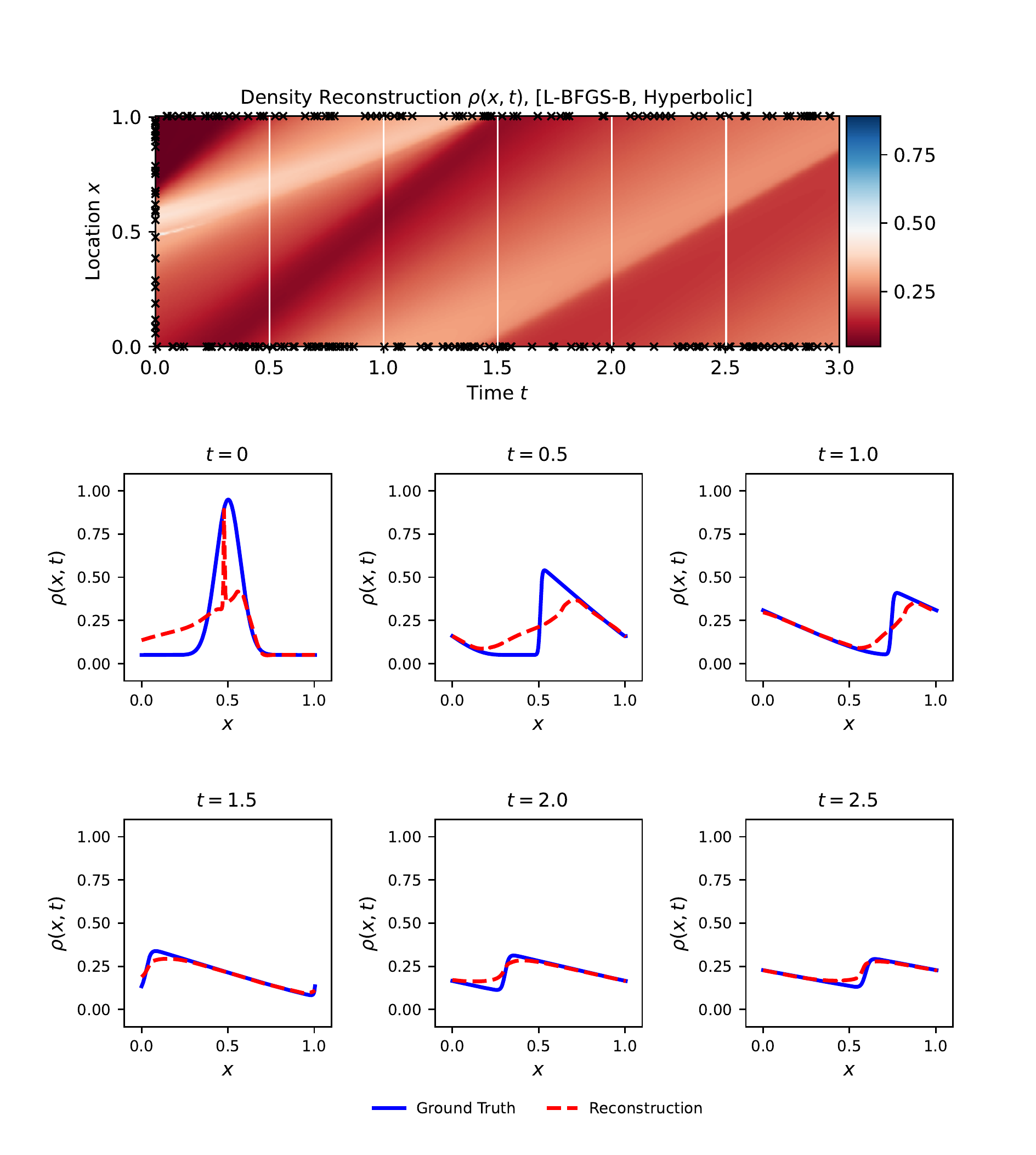}
         \caption{Learning  \textbf{hyperbolic conservation law} using L-BFGS-B optimizer, relative $\mathcal{L}_2$ error: $0.309$}
         \label{fig:reco_bund2}
     \end{subfigure}
     \hfill
     \begin{subfigure}[b]{0.45\textwidth}
         \centering
         \includegraphics[width=3.3in]{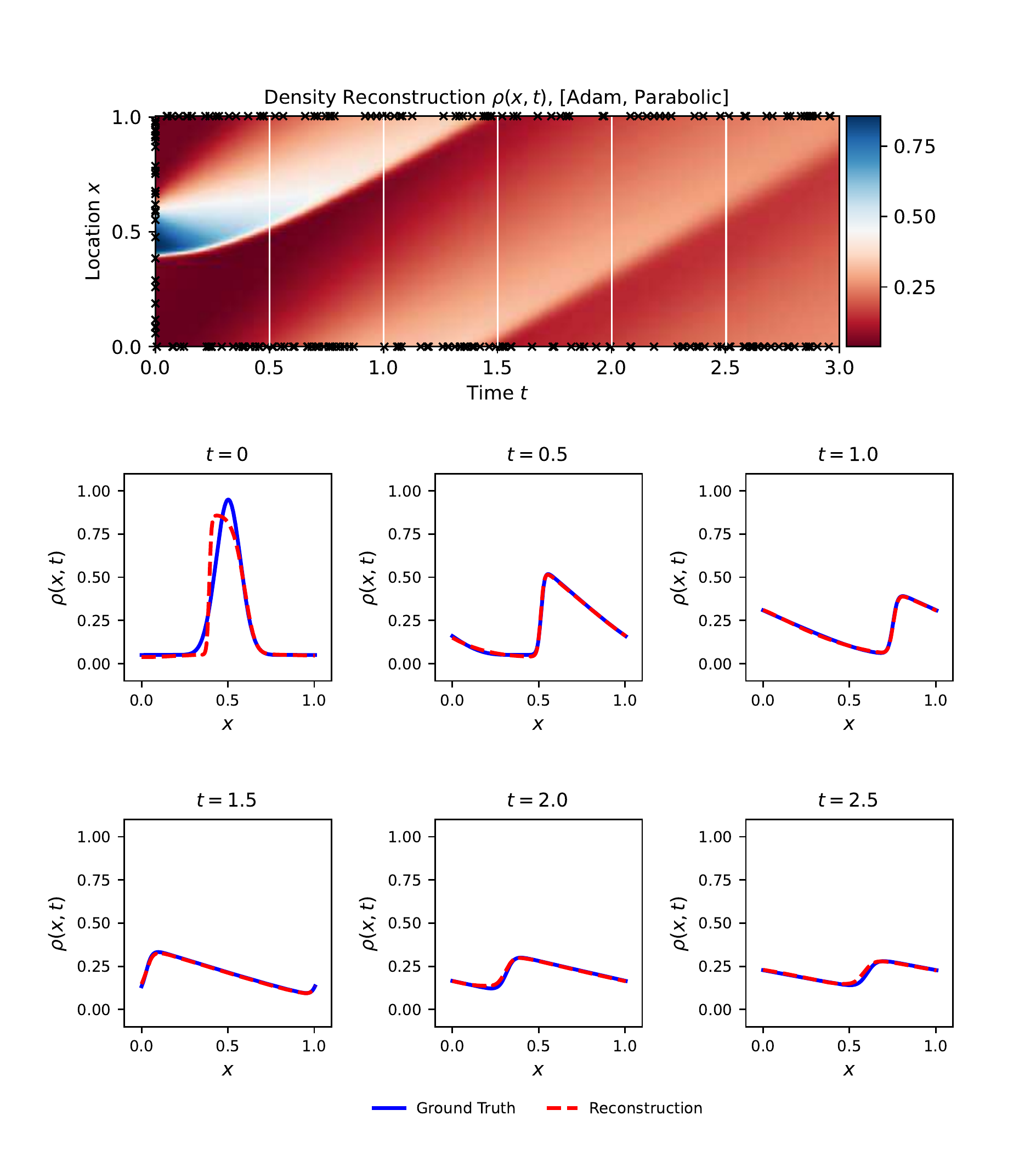}
         \caption{Learning \textbf{parabolic conservation law} using Adam optimizer, relative $\mathcal{L}_2$ error: $0.0630$}
         \label{fig:reco_bund3}
     \end{subfigure}
     \hfill
     \begin{subfigure}[b]{0.45\textwidth}
         \centering
         \includegraphics[width=3.3in]{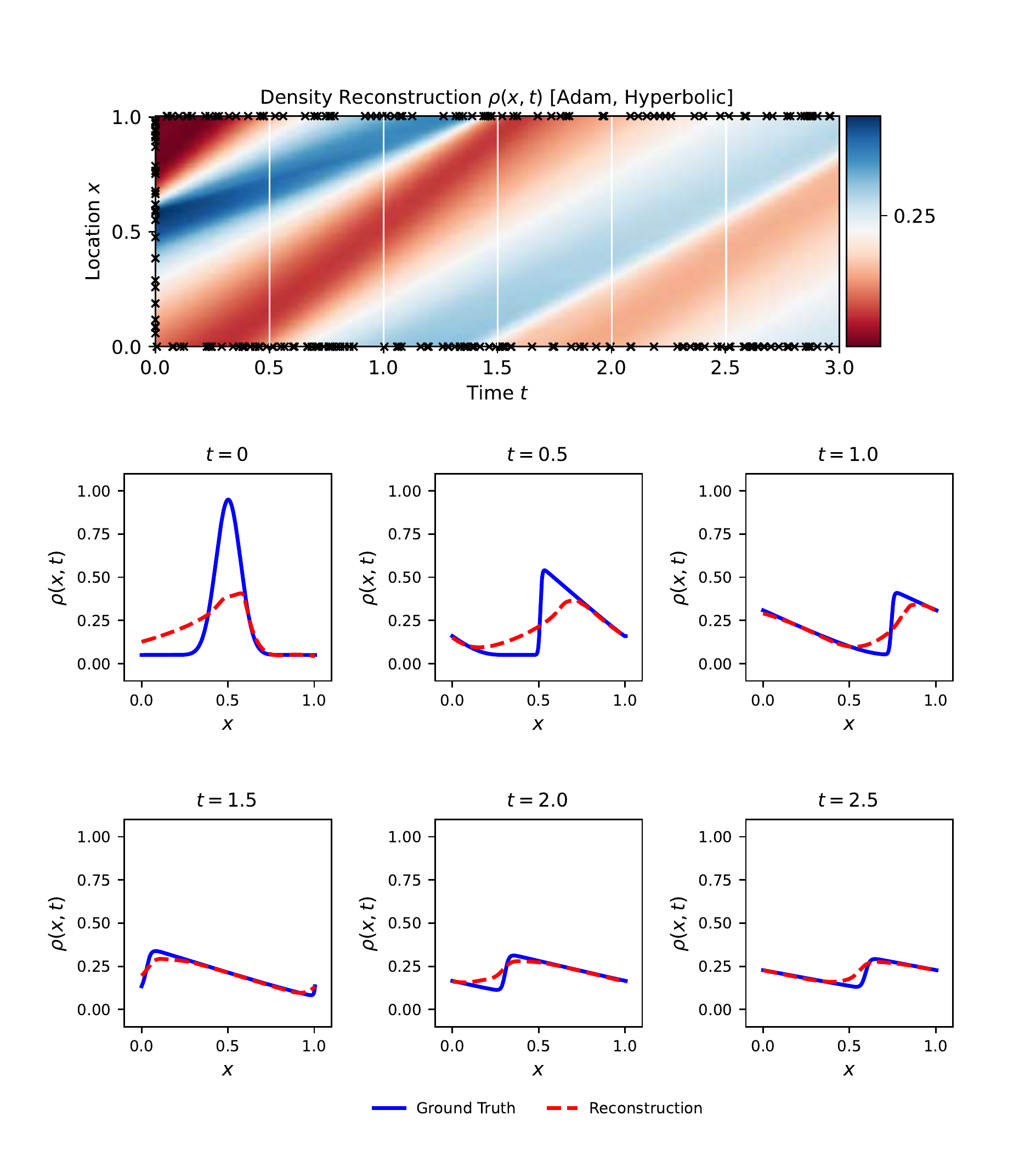}
         \caption{Learning \textbf{hyperbolic conservation law}  using Adam optimizer, relative $\mathcal{L}_2$ error: $0.303$}
         \label{fig:reco_bund4}
     \end{subfigure}
        \caption{Density Reconstruction, Trained with 10\% Initial and Boundary Inputs}
        \label{fig:reco_bund}
\end{figure*}

We first evaluate the PIDL results based only on training inputs about the initial and boundary conditions. We select four levels of available training inputs (10\%, 20\%, 50\%, and 90\% of the total numbers of initial and boundary data), and use both L-BFGS-B and Adam optimizers to reconstruct the hyperbolic LWR PDE and its parabolic variation with the diffusion term. The results are shown in Table~\ref{tab:reco_bund} (best results are shown in \textbf{bold}). The reconstructed density fields, trained with 10\% initial and boundary inputs, are shown in Fig.~\ref{fig:reco_bund}.

\begin{table*}[htbp]
    \caption{Density Reconstruction Results (with Initial and Boundary Inputs), Relative $\mathcal{L}_2$ Error}
    \setlength{\tabcolsep}{3pt}
    \begin{center}
        \begin{tabular}{|p{2.5cm}|p{2cm}|p{2cm}|p{2cm}|p{2cm}|}\hline
            \multirow{2}{*}{\thead[l]{\textbf{Initial and}\\\textbf{Boundary Inputs}\\\textbf{$N_{o1}$}}} 
            & \multicolumn{4}{c|}{\textbf{Relative $\mathcal{L}_2$ Error}} \\\cline{2-5}
            & \thead[l]{\textbf{L-BFGS-B} \\ (parabolic)}
            & \thead[l]{\textbf{L-BFGS-B} \\ (hyperbolic)}
            & \thead[l]{\textbf{Adam} \\ (parabolic)} 
            & \thead[l]{\textbf{Adam} \\ (hyperbolic)} \\\hline
            10\%  & \textbf{1.64e-02} & 3.09e-01 & 6.30e-02 & 3.03e-01\\\hline
            20\%  & \textbf{3.68e-03} & 3.03e-01 & 4.73e-02 & 2.98e-01 \\\hline
            50\%  & \textbf{5.47e-03} & 2.52e-01 & 8.07e-02 & 2.86e-01 \\\hline
            90\%  & \textbf{4.77e-03} & 2.85e-01 & 3.61e-02 & 2.83e-01 \\\hline
        \end{tabular}
        \label{tab:reco_bund}
    \end{center}
\end{table*}

Among all training settings, we observe that PIDL models achieved much higher accuracy (lower relative $\mathcal{L}_2$ error) in reconstructing the parabolic PDE with the diffusion term compared to the ones with the hyperbolic LWR PDE. Trained with the L-BFGS-B optimizer and 10\% boundary and initial observations, the relative $\mathcal{L}_2$ error of PIDL reconstruction is $0.0164$, which is merely \bm{$5.3\%$} of the $0.309$ error of the same metric in reconstructing the hyperbolic LWR PDE.

We also take several snapshots of the reconstruction at time $t = 0, 0.5, 1.0, 1.5, 2.0, 2.5$ for comparison. From Fig.~\ref{fig:reco_bund2}, we notice that PIDL learning the hyperbolic PDE encountered difficulties in reconstructing the density state at locations where the discontinuity of density data occurs. However, in Fig.~\ref{fig:reco_bund1}, equipped with the diffusion term in the parabolic PDE, the reconstruction result is smoothed and closely aligned with the ground truth.

\subsection{RECONSTRUCTION WITH INITIAL, BOUNDARY, AND INTERIOR INPUTS}

\begin{figure*}[htbp]
     \centering
     \begin{subfigure}[b]{0.45\textwidth}
         \centering
         \includegraphics[width=3.3in]{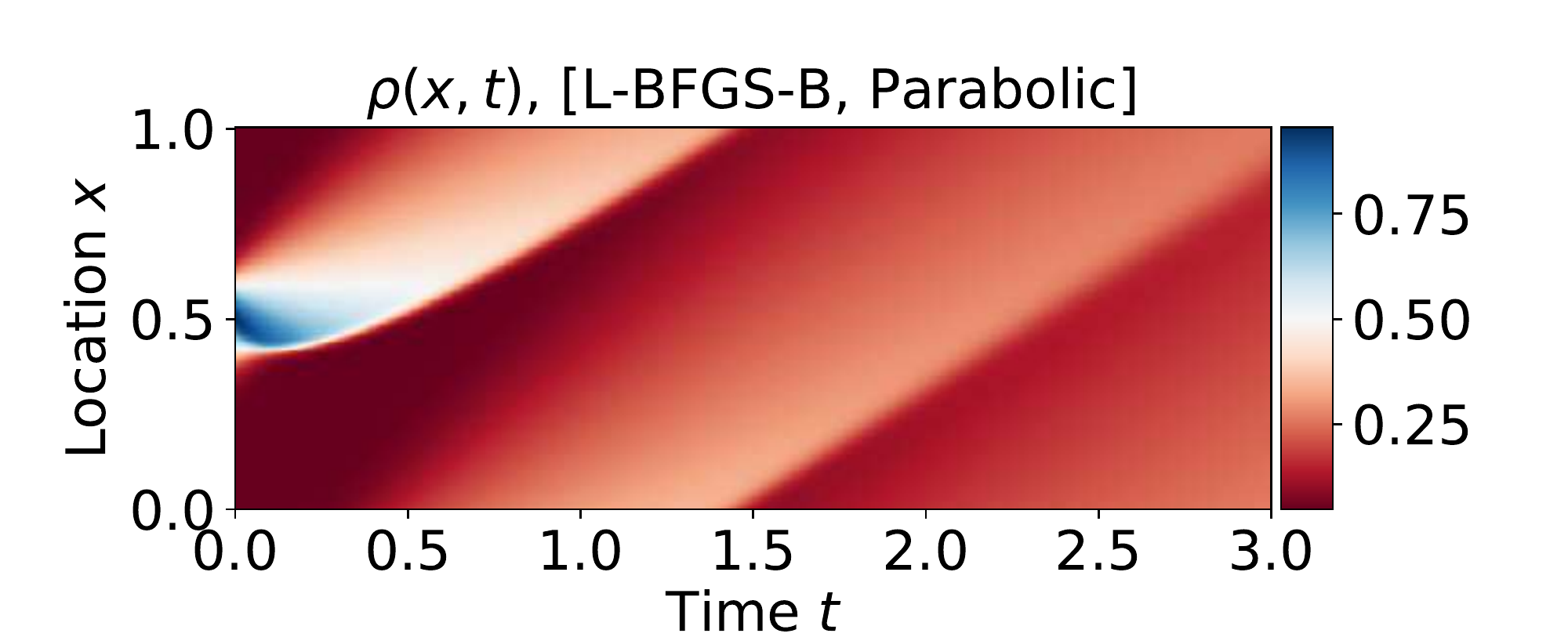}
        \caption{L-BFGS-B with \textbf{parabolic conservation law}, relative $\mathcal{L}_2$ error: $0.00388$}
        \label{fig:reco_inte1}
     \end{subfigure}
     \hfill
     \begin{subfigure}[b]{0.45\textwidth}
         \centering
         \includegraphics[width=3.3in]{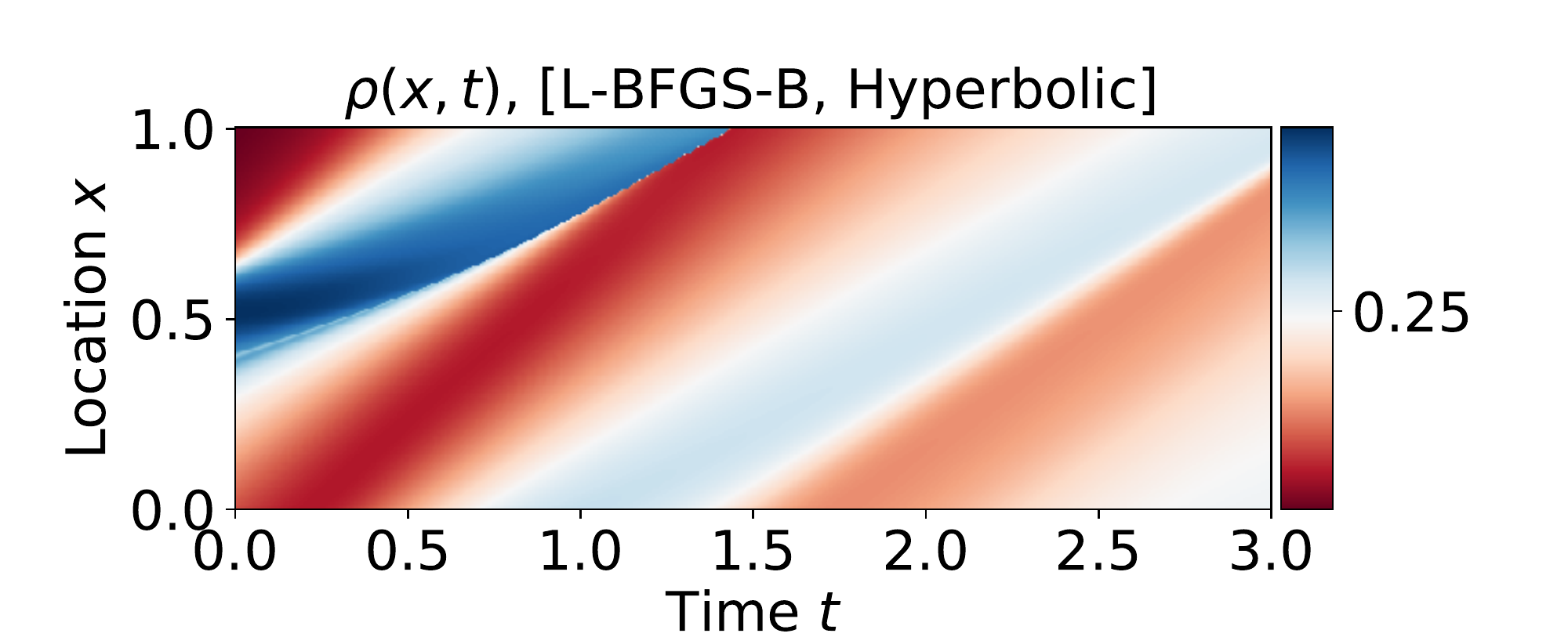}
         \caption{L-BFGS-B with \textbf{hyperbolic conservation law}, relative $\mathcal{L}_2$ error: $0.225$}
         \label{fig:reco_inte2}
     \end{subfigure}
     \hfill
     \begin{subfigure}[b]{0.45\textwidth}
         \centering
         \includegraphics[width=3.3in]{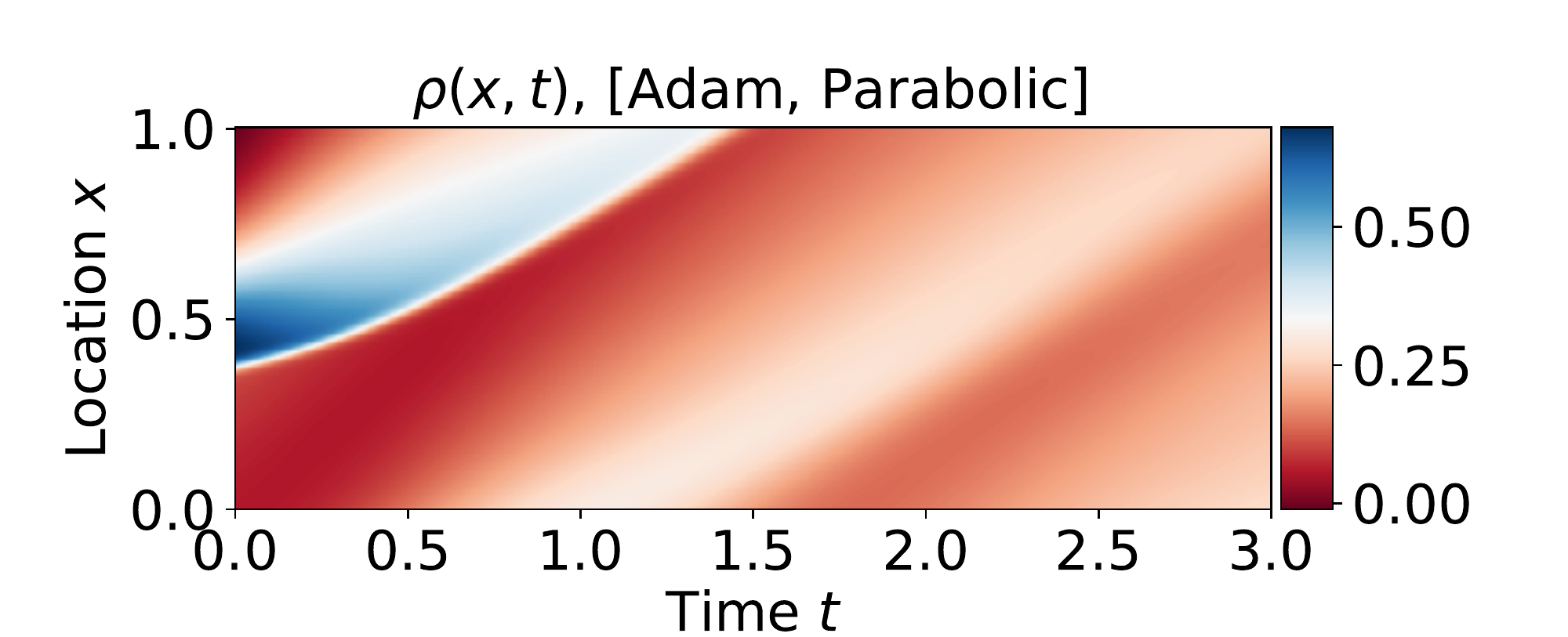}
         \caption{Adam with \textbf{parabolic conservation law}, relative $\mathcal{L}_2$ error: $0.124$}
         \label{fig:reco_inte3}
     \end{subfigure}
     \hfill
     \vspace{1mm}
     \begin{subfigure}[b]{0.45\textwidth}
         \centering
         \includegraphics[width=3.3in]{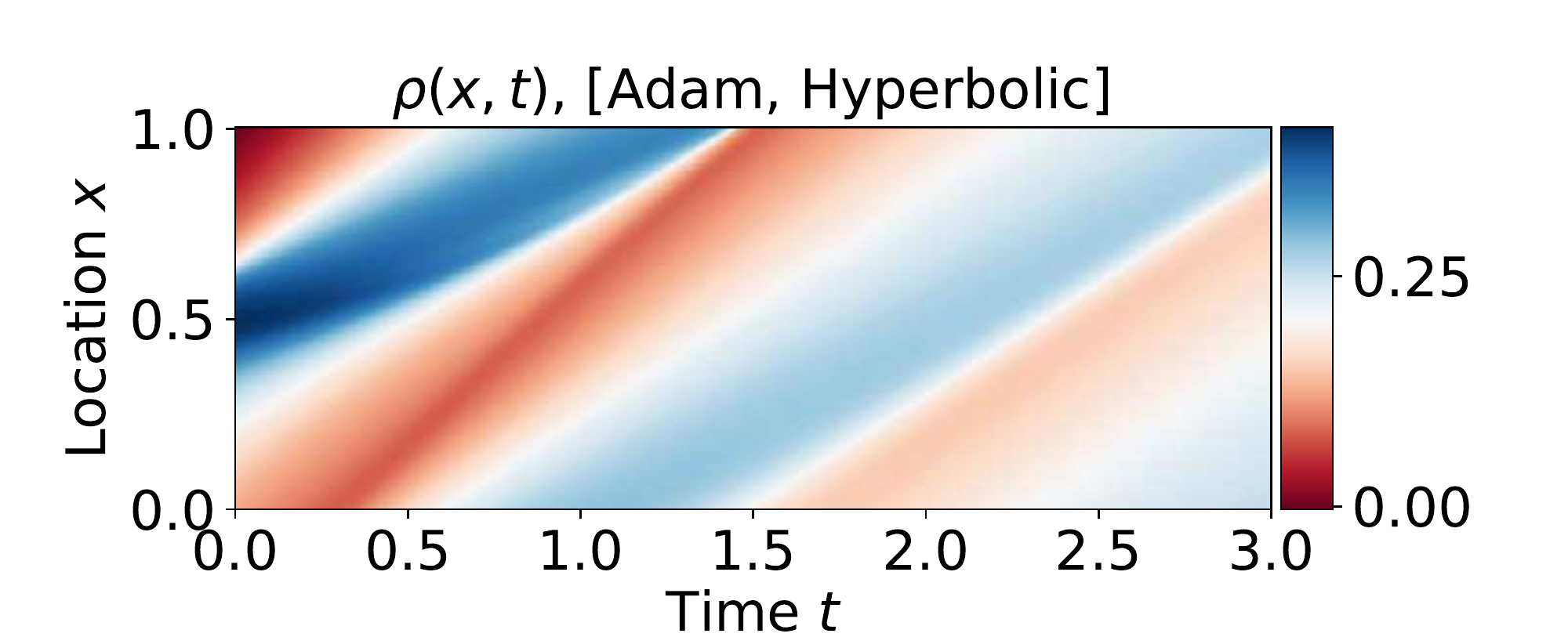}
        \caption{Adam with \textbf{hyperbolic conservation law}, relative $\mathcal{L}_2$ error: $0.247$}
         \label{fig:reco_inte4}
     \end{subfigure}
        \caption{Density Reconstruction, 20\% Initial and Boundary Inputs \& 2\% CV Inputs}
        \label{fig:reco_inte}
\end{figure*}

Subsequently, here we evaluate the reconstruction accuracy under the scenarios in which varying levels of data on the initial and boundary conditions and the interior conditions (CV inputs) are available.  We pick two levels of available inputs $N_{o1}$ on the initial and boundary conditions: (1) $N_{o1} = 108$ inputs, representing $5\%$ of initial and boundary data; and (2) $N_{o1} = 432$ inputs, representing $20\%$ of the available data. For the number of CV inputs $N_{o2}$, we also choose two settings: $N_{o2} = 1146$ and $N_{o2} = 4584$, accounting for $0.5\%$ and $2\%$ of the interior of the density field, respectively. The reconstruction result, measured by the relative $\mathcal{L}_2$ error defined in \eqref{eqn:rl2e} is shown in Table~\ref{tab:reco_inte} (again, best results are tabulated in \textbf{bold}). The reconstruction results with $20\%$ initial and boundary inputs ($N_{o1} = 432$), and $2\%$ CV inputs ($N_{o2} = 4584$) are shown in Fig.~\ref{fig:reco_inte}.

\begin{table*}[htbp]
    \caption{Density Reconstruction Results (with Initial and Boundary Inputs \& CV Inputs), Relative $\mathcal{L}_2$ Error}
    \setlength{\tabcolsep}{3pt}
    \begin{center}
        \begin{tabular}{|p{2.5cm}|p{2.5cm}|p{2cm}|p{2cm}|p{2cm}|p{2cm}|}\hline
            \multirow{2}{*}{\thead[l]{\textbf{Initial and} 
            \\ \textbf{Boundary Inputs} \\
            \textbf{$N_{o1}$}}} & 
            \multirow{2}{*}{\thead[l]{\textbf{CV Inputs} \\ \textbf{$N_{o2}$}}} &
            \multicolumn{4}{c|}{\textbf{Relative $\mathcal{L}_2$ Error}} \\\cline{3-6}
            & & \thead[l]{\textbf{L-BFGS-B} \\ (parabolic)}
            & \thead[l]{\textbf{L-BFGS-B} \\ (hyperbolic)}
            & \thead[l]{\textbf{Adam} \\ (parabolic)} 
            & \thead[l]{\textbf{Adam} \\ (hyperbolic)} \\\hline
            5\%  &  0.5\% & \textbf{8.33e-03} & 2.45e-01 & 7.65e-02 & 2.53e-01\\\hline
            5\%  &  2\% & \textbf{6.05e-03} & 2.37e-01 & 1.41e-01 & 2.66e-01\\\hline
            20\%  & 0.5\% & \textbf{4.50e-03} & 2.23e-01 & 4.29e-02 & 2.48e-01\\\hline
            20\%  &  2\% & \textbf{3.88e-03} & 2.25e-01 & 1.24e-01 & 2.47e-01\\\hline
        \end{tabular}
        \label{tab:reco_inte}
    \end{center}
\end{table*}

Comparing to the results in Table~\ref{tab:reco_bund}, we find the inclusion of CV inputs in the training inputs of PIDL slightly improves the reconstruction accuracy with the hyperbolic PDE. However, the learning performances of PIDL architecture with the parabolic PDE are still far superior. With $20\%$ initial and boundary observations, and $2\%$ CV inputs, the PIDL model with L-BFGS-B optimizer achieves  a relative $\mathcal{L}_2$ error of $0.00388$ for reconstructing the parabolic PDE, which is \bm{$1.72\%$} of the relative $\mathcal{L}_2$ error with the hyperbolic PDE, at $0.225$.

\section{\textcolor{black}{CASE STUDY II - } INSIGHTS FROM FIELD DATA} \label{sec:case2}

In this section, we will further shed light on the topic using field data. We examine the limitation with PIDL and compare it with Lax-Friedrichs' numerical scheme \cite{powers2020numerical} in learning the traffic density using the ``Next Generation SIMulation'' (NGSIM) dataset \cite{ngsim2021}.

\subsection{NGSIM DATASET}

\begin{figure}[htbp]
    \centerline{\includegraphics[width=3.5in]{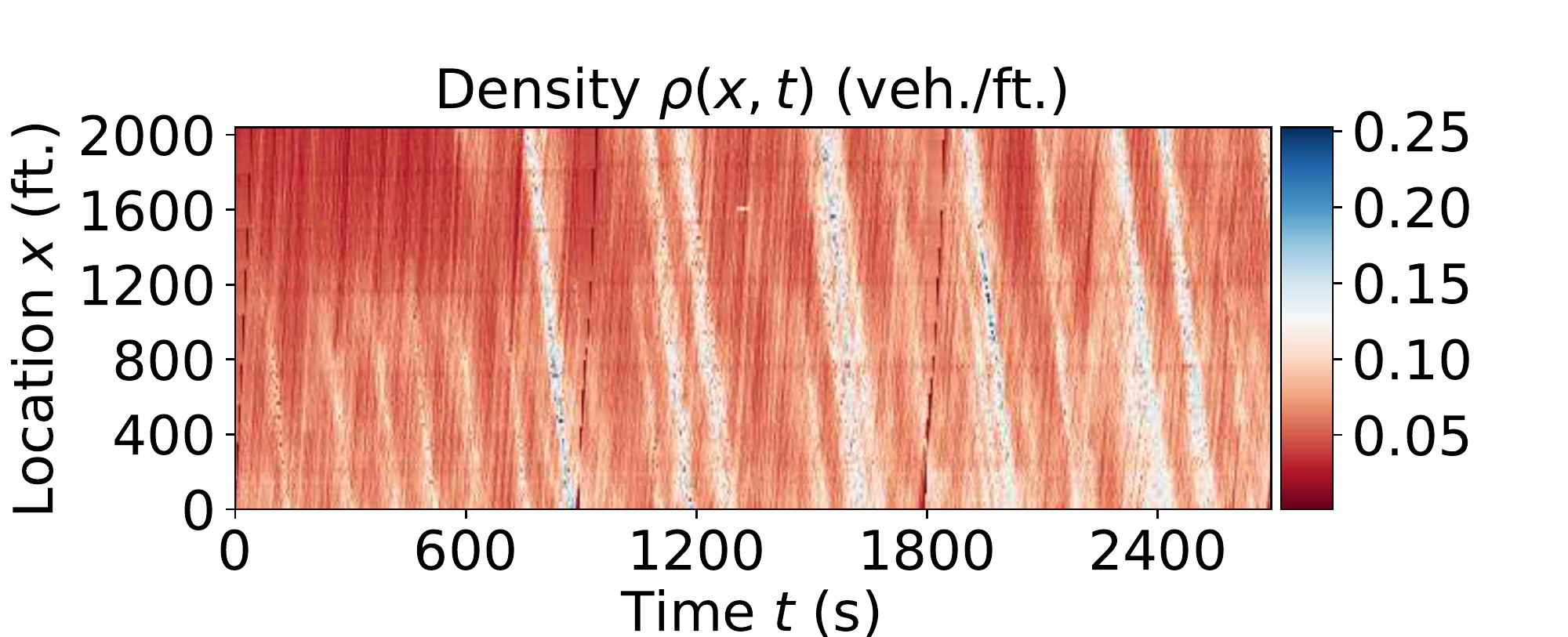}}
    \caption{Vehicle Density on US-101 Highway Segment, between 7:50 am and 8:35 am, NGSIM}
    \label{fig:data_ngsim}
\end{figure}

The NGSIM dataset records traffic conditions using video cameras and processes the traffic state variables such as velocity and vehicle density through vehicle trajectories identified in the video recordings \cite{avila2017}. The vehicle density data used, illustrated in Fig.~\ref{fig:data_ngsim}, contains vehicle density for 45-minute on a 2060-foot segment of $US-101$ freeway. Shockwaves of vehicles stopping due to traffic congestion, which back-propagates in space and forward-propagates in time, can be observed in the plot of vehicle density. \textcolor{black}{This freeway segment has five lanes, one on-ramp and one off-ramp. An additional lane is attached to the freeway between the locations of the on-camp and the off-ramp. The data used was collected from 7:50 a.m. to 8:35 a.m. in Los Angeles, California, on June 15th, 2005. During the first 12 minutes of the data, a free-flow zone is observed in the post-off-ramp area, while the areas before the off-ramp are experiencing stop-and-go wave traffic \cite{avila2020data}.}

\subsection{FIELD DATA RECONSTRUCTION USING PIDL \textcolor{black}{WITH HYPERBOLIC PDE}}

The density dataset is tabulated with spatiotemporal bins of $\Delta x = 20ft$ and $\Delta t = 5s$. At $t = 0$, the initial condition contains 104 data points along the 2060-ft road segment ($x = 0, 20, ..., 2060$). The lower bound condition at $x = 0$ and the upper bound condition at $x = 2060$ each has 540 boundary condition points ($t = 0, 5, ..., 2695$). Together, the initial and boundary condition data of vehicle density $\rho (x, \; t)$ are all used as training inputs of PIDL with the L-BFGS-B optimizer.

\begin{figure}[htbp]
    \centerline{\includegraphics[width=3.5in]{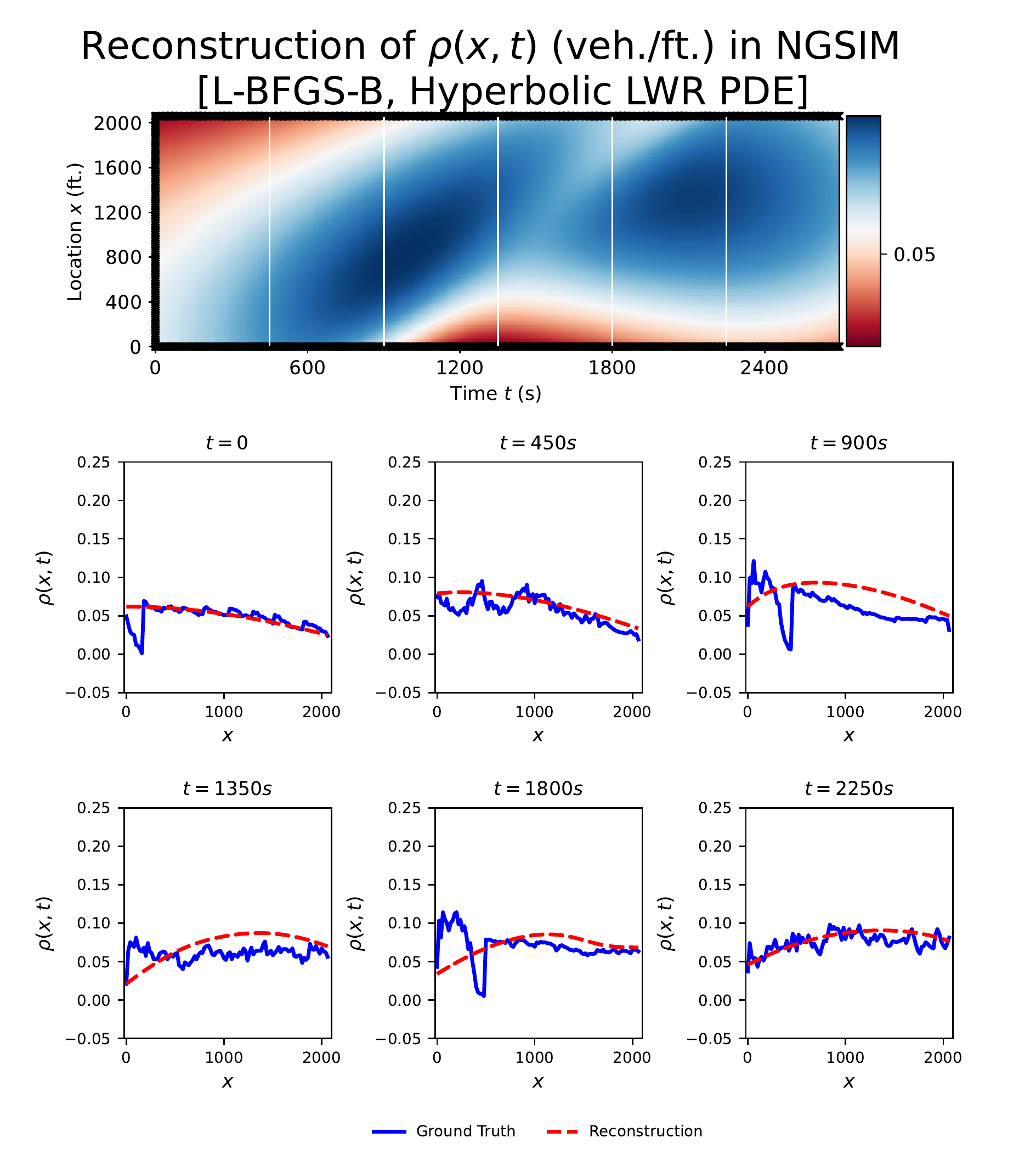}}
    \caption{Reconstruction with PIDL, Relative $\mathcal{L}_2$ Error: $0.345$}
    \label{fig:pidl_ngsim}
\end{figure}

The governing physical equation of the PIDL architecture is the hyperbolic LWR conservation law paired with Greenshield's fundamental diagram. The estimated value of maximum density $\rho_m$ is 0.12 vehicle per foot (sum of all traffic lanes), and the free-flow speed $v_f$ is estimated at 80 feet per second (54.54 miles per hour). The reconstruction result with PIDL is shown in Fig.~\ref{fig:pidl_ngsim}.

The relative $\mathcal{L}_2$ error of PIDL reconstruction is 0.345. From the snapshots of $ t = 0, 450, 900, 1350, 1800, 2250$, PIDL tries to mimic the evolution of the density field between the lower boundary $x = 0$ and the upper boundary $x = 2060$; however, it cannot overcome the challenge in learning the stochastic perturbation of the density state. It is also evident in the reconstruction plot of $\rho(x \; t)$ that the PIDL architecture overly generalizes the output and fails to capture any traffic patterns, such as the shockwaves present in the dataset.

\textcolor{black}{\subsection{FIELD DATA RECONSTRUCTION USING PIDL WITH PARABOLIC PDE}}

\begin{figure}[htbp]
    \centerline{\includegraphics[width=3.5in]{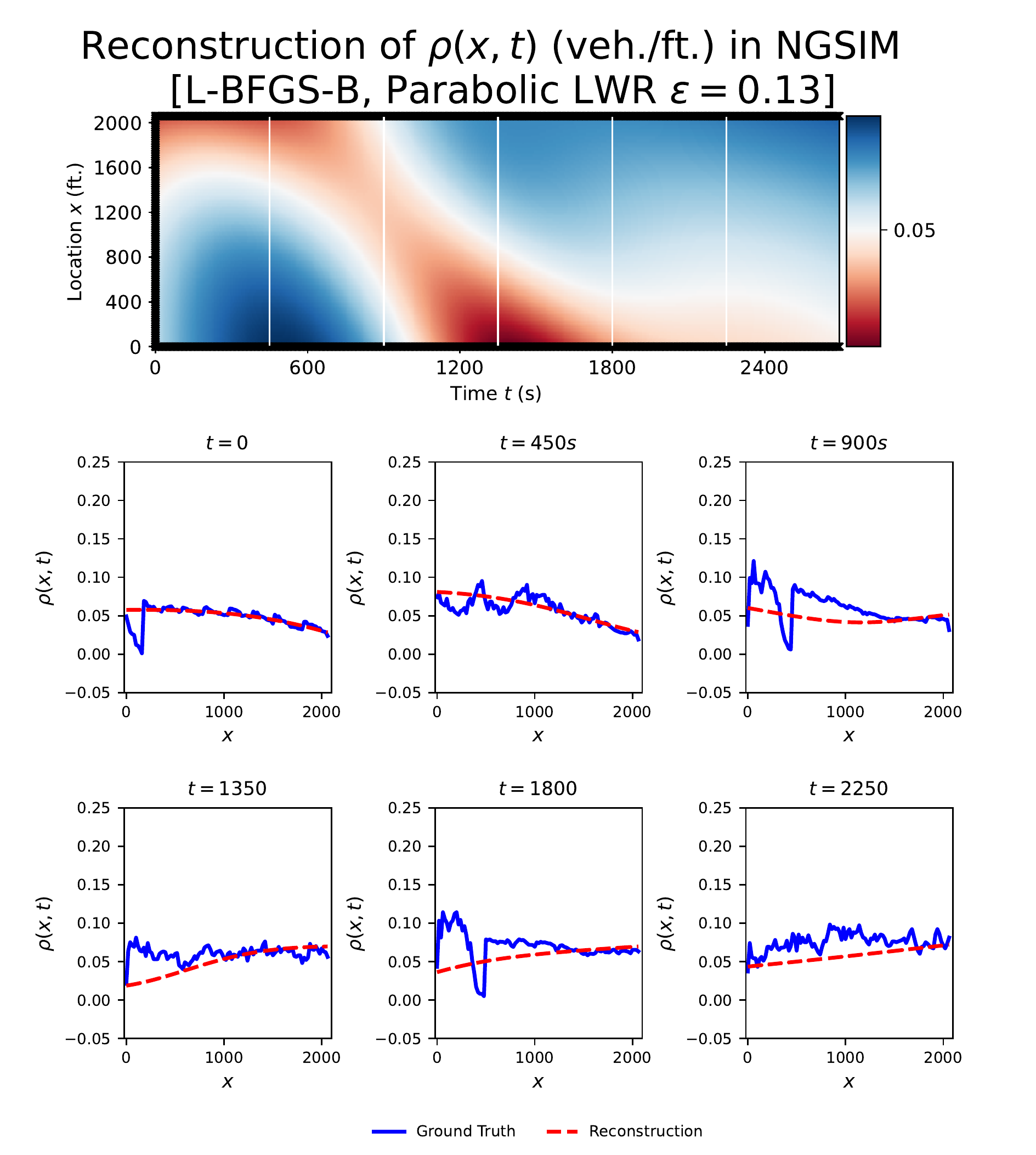}}
    \caption{\textcolor{black}{Reconstruction with PIDL, $\epsilon = 0.13$, Relative $\mathcal{L}_2$ Error: $0.319$}}
    \label{fig:pidl_ngsim_para}
\end{figure}

\textcolor{black}{
 For NGSIM US-101 data, the diffusion coefficient $\epsilon$ in Eqn.~\eqref{eqn:lwrd_para} is estimated to be $0.13$ for regular motor vehicles \cite{xu2020driver}. We select a range of values at ${0.05, 0.1, 0.13, 0.15, 0.20}$ for $\epsilon$ and reconstruct the density dataset using the LWR PDE with the addition of the diffusion term as the physics of PIDL. The reconstruction result with $\epsilon = 0.13$ is shown in Fig.~\ref{fig:pidl_ngsim_para}.
}

\textcolor{black}{
We observe that with the realistic NGSIM dataset, the addition of the diffusion term only slightly improves the accuracy, landing a relative $\mathcal{L}_2$ error at 0.319. Similar to the troubles the PIDL neural network with hyperbolic LWR PDE encountered, the stochastic nature of traffic disturbance overcomes the PIDL's ability to accurately capture the traffic state, with \textit{only} boundary and initial observations. Our previous work \cite{huang2022physics} demonstrates that the inclusion of Lagrangian observations, such as CV data, will significantly improve PIDL performances in this case. 
}

\textcolor{black}{
We also conduct the sensitivity analysis on the diffusion coefficient $\epsilon$, and the results are presented in Table~\ref{tab:epsi}. With varying values of the diffusion coefficient $\epsilon$, the conclusion from this analysis is unwavering: although the diffusion term smooths out the reconstruction in areas where state discontinuity is present, it cannot accurately estimate the traffic state with \textit{only} initial and boundary observations.
}

\begin{table}[htbp]
    \caption{\textcolor{black}{Relative $\mathcal{L}_2$ Error with Choices of Diffusion Coefficient $\epsilon$}}
    \setlength{\tabcolsep}{3pt}
    \begin{center}
        \begin{tabular}{|p{2.5cm}|p{4.0cm}|}\hline
            \textbf{Value of $\epsilon$}  &  \textbf{Relative $\mathcal{L}_2$ Error}\\\hline
            0.05  &  3.44e-01  \\\hline
            0.10  &  3.23e-01  \\\hline
            0.13  &  3.19e-01  \\\hline
            \textbf{0.15}  &  \textbf{3.10e-01}  \\\hline
            0.20  &  3.24e-01  \\\hline
        \end{tabular}
        \label{tab:epsi}
    \end{center}
\end{table}

\subsection{RECONSTRUCTION USING LAX-FRIEDRICHS' NUMERICAL SCHEME}

\begin{figure}[htbp]
    \centerline{\includegraphics[width=3.5in]{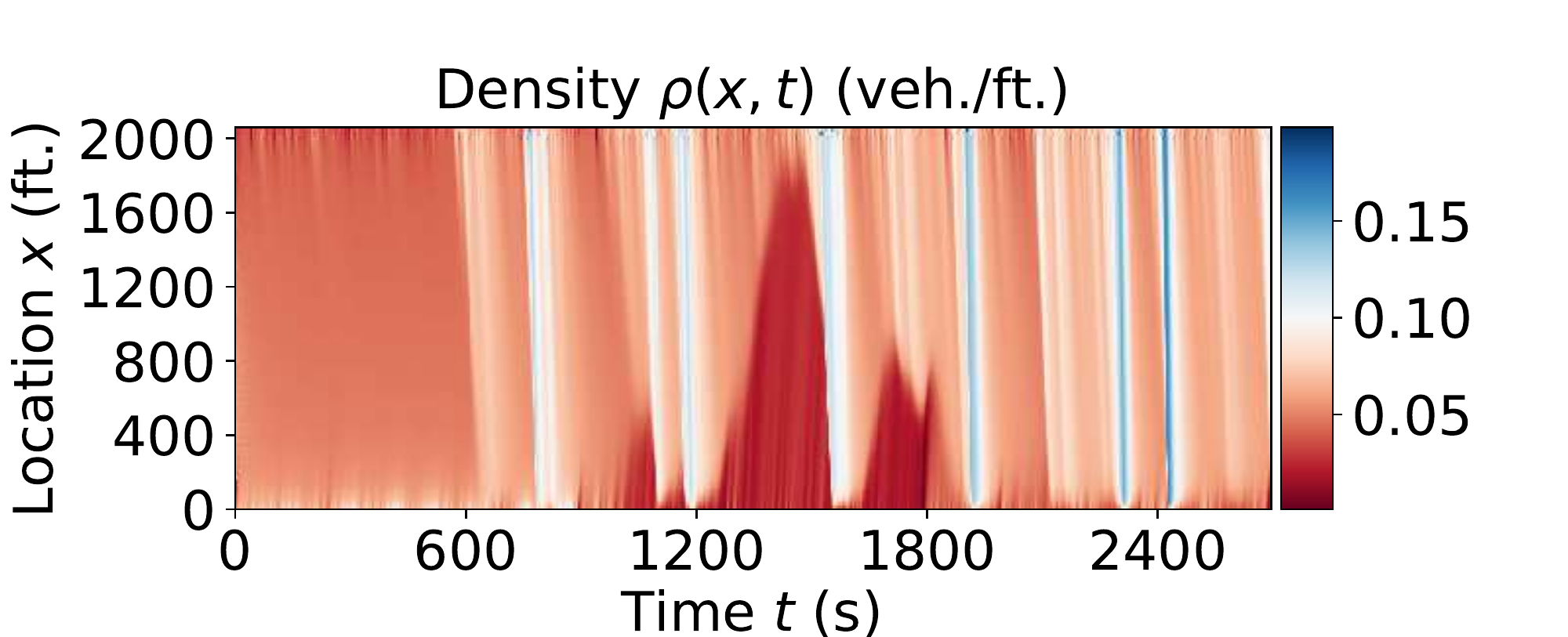}}
    \caption{Reconstruction with Lax-Friedrichs' Numerical Scheme, Relative $\mathcal{L}_2$ Error: $0.231$}
    \label{fig:lax_ngsim}
\end{figure}

The vehicle density dataset from NGSIM can also be reconstructed by using the Lax-Friedrichs' differencing method \cite{leveque1992numerical} with the complete initial and boundary conditions. The reconstruction is pictured in Fig.~\ref{fig:lax_ngsim}.

Along with a smaller relative $\mathcal{L}_2$ error, the most significant improvement in the reconstruction result by the Lax-Friedrichs' method is the capability to capture and rebuilds the pattern of shockwaves in the dataset, based only on the inputs from the initial and boundary condition. The advantage of the numerical scheme resides in the ability to propagate the discontinuity in the density field based on the method of characteristics, while the PIDL architecture struggles with the reconstruction task.

\section{\textcolor{black}{DISCUSSION ON RESULTS}} \label{sec:disc}

\textcolor{black}{
Recent examinations have elucidated the training limitations and drawbacks of data representation in PIDL. In several instances, unstable convergence occurs in the gradient-descent-based PIDL training, especially when the underlying PDE solution has high-frequency features \cite{zhu2019physics}. This pathological behavior observed in PIDL training is due to the multi-scale interactions between the cost terms in optimizing the neural network cost \cite{wang2021understanding}. It leads to stiffness in the gradient flow dynamics, ultimately inducing a severe constraint on the learning rate and adding detriment to the stability of the training process. PIDL, which often deploys fully-connected hidden layers, faces the challenge termed ``spectral bias'' that cannot reasonably assimilate a nonlinear hyperbolic PDE when its solution involves shocks in the field of data \cite{wang2022and}. For a one-dimensional hyperbolic PDE with a non-convex flux function, its analytical solution can be depicted by a simple piecewise continuous function, and the stability of its solution can be significantly improved by adding a diffusion term to the inherent PDE. With smoothing around the shock by the diffusion, the neural network can recuperate the actual scale and location of the shock, solves the PDE in its parabolic form, and this leads to precise approximation results \cite{fuks2020limitations}.
}

\textcolor{black}{
\textbf{Suggestions on improvement:}
As we observe the physics-informed deep learning neural networks encounter struggles with approximating the vehicle density where localized non-linear discontinuity exists in the data, artificial dissipation should be added in order to improve the learning result of the hyperbolic conservation law \cite{fraces2020physics, gasmi2021physics}, and increased number of observations along the shock trajectories should be included in the training of the neural network \cite{rodriguez2022physics}.
}

\textcolor{black}{
Common practices assume the coefficient $\epsilon$ to be zero when fitting a realistic traffic dataset with LWR conservation law \cite{fan2013data, shi2021physics}, leaving out the diffusion effect. From Fig.~\ref{fig:reco_bund1} and Fig.~\ref{fig:reco_bund2} in Section~\ref{sec:case}, the inclusion of the diffusion term significantly improved the reconstruction accuracy of a PIDL neural network. The deficiency of PIDL with hyperbolic LWR PDE is not rooted in the learning architecture of the neural network or the hyperparameter selection \cite{fuks2020limitations}. The diffusion term enhances the stability of the gradient optimization process in training the neural network around the areas with state discontinuity and shockwaves.
}

\textcolor{black}{
Based on the results in Section~\ref{sec:case2}, PIDL with the parabolic variant of LWR PDE cannot overcome the random perturbation in the traffic dataset and accurately estimate the traffic density based on pure observation of the initial and boundary conditions. Our previous work suggests the benefit of including Lagrangian observation in this setting for the task of TSE \cite{huang2022physics}.
} 

\textcolor{black}
{In practice, the task of TSE involves realistic traffic data in which a diffusion term is inherent - drivers will gradually slow down the speed of their vehicles in participation of congestion or when a slowdown is visually perceivable. Therefore when adopting physics-informed deep learning for TSE, this nature of smoothness around shockwaves should be considered as part of the ``physics'', which illustrates the underlying relationship between traffic states. Recall the diffusion term in the parabolic form of LWR PDE is weighted by the parameter $\epsilon$. The value of $\epsilon$ is associated with drivers' behavior (reaction time to press the brake, for example) and should be tuned according to the traffic dataset before plugging in the conservation equation for the realistic reconstruction of the traffic data. Other approaches that transportation planners and agency practitioners can adopt is to include Lagrangian sensing data \cite{huang2020physics, huang2022physics}, increasing the number of observation samples in the shockwave areas \cite{rodriguez2022physics}, and domain decomposition and projection onto the space of high-order polynomials \cite{kharazmi2021hp}.  
}

\section{CONCLUSION} \label{sec:conc}

In this work, we exhibited the difficulties of training a physics-informed deep learning (PIDL) neural network to reconstruct a certain type of partial differential equation (PDE) - the hyperbolic PDE for which a strong solution cannot be obtained. The non-smooth weak solution to conservation law-based traffic flow models (such as the LWR PDE) causes PIDL failure in capturing the scale and location of the discontinuity. Through the case study, we showcased the stark differences between the learning result using PIDL with the first order hyperbolic LWR PDE and its parabolic counterpart, in which the additional diffusion term secures a strong solution and leads to pinpoint approximation. \textcolor{black}{When the PDE solution contains multi-scale features of high-frequency terms, it oftentimes causes severeness in calculating the gradients in the fully connected learning structure of a PIDL neural network \cite{wang2022and}, making the optimization process unstable and ultimately leads to inaccurate predictions \cite{raissi2018deep, zhu2019physics}. We observe that the deep learning neural network fails to approximate the nonlinear relationship of a hyperbolic PDE in areas where shockwaves are present, whereas the diffusion term in the parabolic PDE ensures improved data estimation in these areas and thus ameliorates the reconstruction result.} Besides reconstructing using only the initial and boundary conditions, we list the possible sampling choices of traffic flow and use a diverse combination of Eulerian and Lagrangian data to ensure the reliability and validity of the results. Moreover, with the field data from the NGSIM traffic dataset, we further highlight the limitation of PIDL in the presence of shockwaves. Future work includes analysis of the cost evolution (over time) in reconstructing hyperbolic and parabolic PDEs to understand the interaction between the cost terms of the PIDL neural network and interpret its pathological behavior in learning conservation law-based models.

\section*{ACKNOWLEDGMENT}
The authors wish to thank Dr. Rongye Shi at Columbia University for providing the ring road data for the case study, Dr. Animesh Biswas at the University of Nebraska at Lincoln for the LWR reconstruction of the data, and Dr. Pushkin Kachroo at the University of Nevada Las Vegas for the suggestions and discussions on the topic.

\bibliographystyle{unsrt}
\bibliography{limitation} 
\end{document}